\documentclass{article}

\PassOptionsToPackage{numbers}{natbib}
\usepackage{enumitem}
\usepackage[main, final]{neurips_2026}

\usepackage[utf8]{inputenc} %
\usepackage[T1]{fontenc}    %
\usepackage{hyperref}       %
\usepackage{url}            %
\usepackage{booktabs}       %
\usepackage{amsfonts}       %
\usepackage{amssymb}        %
\usepackage{nicefrac}       %
\usepackage{microtype}      %
\usepackage{xcolor}         %

\definecolor{canonblue}{HTML}{60A5FA}   %
\definecolor{canonorange}{HTML}{FB923C} %
\definecolor{canongreen}{HTML}{4ADE80}  %

\usepackage{graphicx}
\usepackage[font=small]{caption} %
\usepackage{subcaption}  %
\usepackage{wrapfig}     %
\usepackage{comment}
\usepackage[most]{tcolorbox} %
\usepackage{cleveref}    %
\usepackage{xspace}
\usepackage{xparse}

\newif\ifvenueicml
\newif\ifvenueneurips
\newif\ifvenuearxiv
\venuearxivtrue   %

\providecolor{canonblue}{HTML}{60A5FA}   %
\providecolor{canonorange}{HTML}{FB923C} %
\providecolor{canongreen}{HTML}{4ADE80}  %

\usepackage[normalem]{ulem}

\newtcolorbox{promptbox}[1][]{
  enhanced,
  breakable,
  width=\dimexpr\linewidth-5pt\relax,
  colback=gray!15,
  colframe=gray!55!black,
  coltext=black,
  boxrule=0.4pt,
  arc=2pt,
  left=8pt, right=8pt, top=6pt, bottom=6pt,
  fontupper=\ttfamily\small,
  fonttitle=\bfseries\sffamily\small,
  coltitle=white,
  colbacktitle=gray!55!black,
  attach boxed title to top left={xshift=8pt, yshift=-2pt},
  boxed title style={colframe=gray!55!black, sharp corners, boxrule=0pt},
  #1,
}

\newtcolorbox{claim}{
    colback=gray!5,
    colframe=gray!60,
    boxrule=0.5pt,
    arc=2pt,
    left=6pt,
    right=6pt,
    top=2pt,
    bottom=2pt,
    before skip=4pt,
    after skip=4pt
}

\newcommand{\Orig}{\ifmmode\mathrm{Orig}\else Orig\fi\xspace}
\newcommand{\Self}{\ifmmode\mathrm{Self}\else Self\fi\xspace}

\newcommand{\M}{\ensuremath{\mathcal{M}}\xspace}
\newcommand{\Mzero}{\ensuremath{\mathcal{M}_0}\xspace}
\newcommand{\Mreg}{\ensuremath{\mathcal{M}_{\mathrm{reg}}}\xspace}

\newcommand{\Mregk}[1]{\ensuremath{\mathcal{M}_{\mathrm{reg}}^{(#1)}} \xspace}
\newcommand{\Munreg}{\ensuremath{\mathcal{M}_{\mathrm{unreg}}}\xspace}

\newcommand{\MLlama}{\ensuremath{\mathcal{M}_{\mathrm{Llama}}}\xspace}
\newcommand{\Mmix}
{\ensuremath{\mathcal{M}_{\mathrm{mix}}}\xspace}

\newcommand{\BB}[1]{\ensuremath{B(#1)}\xspace}
\newcommand{\EE}[1]{\ensuremath{E(#1)}\xspace}

\newcommand{\metaq}{\ensuremath{q}\xspace}
\newcommand{\metae}{\ensuremath{e}\xspace}
\newcommand{\metaehat}{\ensuremath{\hat{e}}\xspace}

\newcommand{\Mtrain}[1]{\ensuremath{\mathcal{M}[{#1}]}\xspace}

\newcommand{\Jtrain}{\ensuremath{J_{\mathrm{train}}}\xspace}
\newcommand{\Jtest}{\ensuremath{J_{\mathrm{test}}}\xspace}

\newcommand{\runningexamplepanel}[5]{%
\begin{minipage}[t]{0.48\linewidth}
\setlength{\fboxsep}{6pt}%
\fcolorbox{black!25}{gray!4}{%
\begin{minipage}{0.92\linewidth}
\raggedright
\textbf{#1}\\[-1pt]
\hrule
\vspace{4pt}
{\ttfamily #2}
\vspace{4pt}

\hrule
\vspace{4pt}
\textbf{Model answer:}~#3
\ifx\relax#4\relax\else\\{\footnotesize #4}\fi
\ifx\relax#5\relax\else\\{\footnotesize #5}\fi
\vspace{2pt}
\end{minipage}}%
\end{minipage}%
}

\title{
Introspective Coupling: Self-Explanation Training Tracks Behavioral Change Despite Fixed Supervision
}

\author{%
  Zifan Carl Guo\thanks{Correspondence to \texttt{carlguo@mit.edu}. Code: \url{https://github.com/carlguo866/introspective-coupling}.} \quad Laura Ruis \quad Jacob Andreas \quad Belinda Z. Li \\
  MIT EECS
}

\begin{document}

\venuearxivfalse \venueneuripstrue

\maketitle

\begin{abstract}
When does training language models (LMs) to generate explanations of their predictions yield faithful introspection, rather than superficial imitation? 
We study LMs trained to explain which features of their inputs influenced their behavior, using models' counterfactual behavior on modified inputs as supervision.
Surprisingly, we find that LMs trained on fixed counterfactual explanations derived from their earlier checkpoints, or even from behaviorally similar models in different families, %
frequently produce explanations more faithful to \textit{their own current behaviors} than to those of their training targets.
This ``introspective'' coupling between LM explanations and behaviors occurs when 
training explanations remain sufficiently correlated with current behaviors over the course of training, even as behaviors themselves shift.
We also show that introspective coupling tracks %
behavior shifts: when explanation training is provided concurrently with other post-training objectives, explanations change to match without requiring updated supervision. This phenomenon appears in multiple tasks, including sycophancy and refusal, and is robust to label noise.
Overall, our results show that even fixed datasets of counterfactual explanations can provide scalable and generalizable post-training signal for introspection.
\end{abstract}

\begin{figure*}[t] %
    \centering
    \includegraphics[width=0.9\linewidth]{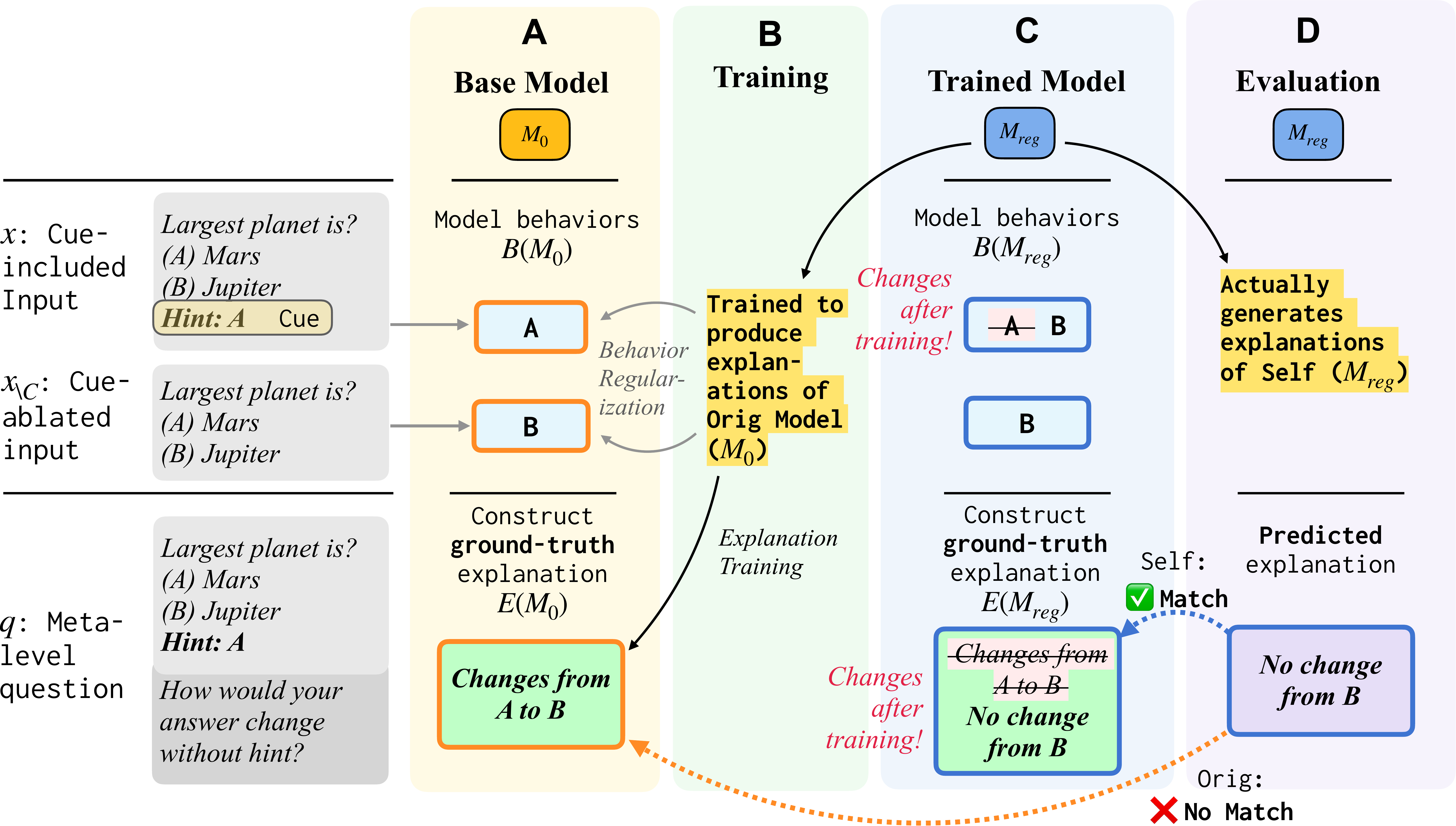}
    \caption{\textbf{Method overview.}
    (A) We sample behaviors $\BB{\Mzero}$ from the base model $\Mzero$ on inputs $x,x_{\setminus C}$ and construct explanation labels $\EE{\Mzero}$.
    (B) We fine-tune $\Mzero$ on these explanations, yielding $\Mreg$.
    (C) During training, $\Mreg$'s behavior drifts to distribution $\BB{\Mreg}$, so its corresponding explanations shift to $\EE{\Mreg}$.
    (D) $\Mreg$’s output explanations track this change: they better match explanations of $\Mreg$’s own behavior (\Self) despite training on explanations of $\Mzero$ (\Orig). We call this Self $>$ Orig \textit{introspective coupling}.
     }
    \label{fig:overview}
    \vspace{-1.5em}
\end{figure*}

\section{Introduction}
\label{sec:introduction}

Methods for training language models (LMs) to faithfully output reasons behind their predictions and decisions offer a promising avenue for understanding LM behavior. Such \textit{introspective} abilities could enable monitoring, debugging, and reasoning about behavioral changes under distribution shift. Past work has studied many forms of self-explanation, including explanations of features~\citep{pan2024latentqa,karvonen2026activationoraclestrainingevaluating},  circuits~\citep{li2025traininglanguagemodelsexplain,lindsey2025emergent}, and behavioral traits~\citep{binder2025looking,plunkett2025selfinterpretabilityllmscomplexinternal,hase2026counterfactual}. Yet a central challenge is ensuring that LM-generated explanations track an LM's own behavior, rather than merely imitating plausible patterns. 

Following past works studying self-explanation in LMs~\citep{hase2026counterfactual,li2025traininglanguagemodelsexplain,turpin2023language}, we adopt a counterfactual simulability view of explanation, in which explanations are judged to be faithful if they identify parts of the input (e.g. clues on the answer) that influence model decision-making, and are evaluated by testing whether model behavior changes when (only) identified features are perturbed (\Cref{fig:overview}). 

While self-explanation ability %
may emerge from sufficiently large models~\citep{lindsey2025emergent}, out-of-the-box models are not guaranteed to generate faithful explanations~\citep{turpin2023language,madsen2024self,chen2025reasoningmodelsdontsay,barez2025cot}. 
This has motivated a line of work that explicitly trains models to generate explanations. A common recipe is supervised fine-tuning (SFT) on a static set of explanation labels derived from an initial snapshot of the model being explained~\citep{binder2025looking,li2025do}.
This approach
has an underexamined limitation: 
as an LM is trained to generate explanations, its behavior on non-explanation inputs may drift, so
the model being trained learns to explain the behavior of the earlier checkpoint, rather than its own, current behavior. 
However, for a self-explanation to be faithful, it must track the model's current behavior rather than a stale set of behavior before explanation training.

In this paper, we make the surprising observation that training models to generate explanations of their earlier behavior, while regularizing behaviors themselves toward an earlier checkpoint,  
\emph{in fact causes them to explain their current behavior more faithfully than the checkpoint from which the explanations were derived} (\Cref{fig:overview}).
On test data, regularized explanation training reduces drift relative to unregularized fine-tuning, but does not eliminate it entirely. A trained model's explanations track this residual drift, despite never receiving supervision derived from its current, on-policy behavior.
This effect disappears without regularization.

More formally, we call this effect \textit{Self > Orig}, where:
\begin{claim}
A model demonstrates
\textbf{Self > Orig} if its explanations more faithfully predict its current behavior than the original behavior used to generate its training labels.
\end{claim}

We interpret Self > Orig as evidence of \textit{introspective coupling}:
\begin{claim}
\textbf{Introspective Coupling}: a phenomenon where a model learns to couple its explanations to its own current behavior rather than merely reproducing static explanation training targets. 
\end{claim}

We show that \textbf{Self > Orig} is robust across three tasks (\Cref{sec:self_vs_original}): two sycophancy datasets (Hint-MMLU~\citep{chen2025reasoningmodelsdontsay} and AITA~\citep{cheng2026elephant}), and a refusal dataset comprising a mix of FalseReject~\citep{zhang2025falsereject} and WildJailbreak~\citep{jiang2024wildteaming}.
In each case, regularized training produces models whose
explanations are more faithful to their \textit{current} behavior than to the behavior of the original model that generated their training labels.

We further show that introspective coupling
leaves a mechanistic fingerprint, coupling not only behaviors and explanations but the internal representations that produce them 
(\Cref{sec:self_vs_original:interpretability}). 

We characterize \textit{when} introspective coupling emerges (\Cref{sec:when}), finding that 
explanation training data must remain similar to a model's \textit{current, online} behavior---even if explanations diverge from the model's \textit{original} behavior by up to 50\%. This holds promise for scaling explanation training: not only do we not need to regenerate explanation labels over the course of training, but we may also be able to reuse labels from a sufficiently similar separate model and \textit{still induce} introspective coupling.

Finally, we demonstrate that introspective coupling remains useful when a model acquires new behavior beyond the original explanation labels (\Cref{sec:generalization}). We train models on auxiliary data that alters their underlying behavior distribution, either directly or indirectly, and find that explanation tracks those changes.
This is promising for integrating introspection training into standard post-training pipelines: models' explanations can track behavioral drifts induced by different supervisions.\footnote{We refer to ``introspection'' in the context of the operationalization above where models exhibit \textbf{behavioral} introspection that matches its own object-level outputs. We emphasize that there is an ongoing debate about the specific definition of model introspection and do not make claims that models have metacognitive awareness or complete access to all internal computations.}

\section{Methods}
\label{sec:methods}

\subsection{(Counterfactual) Explanation Construction}
\label{sec:methods:framework}
Following past work, we focus on counterfactual explanations: how behaviors change under an edit to its input (e.g. removal of a hint)~\citep{li2025traininglanguagemodelsexplain,hase2026counterfactual}. 
These allow us to identify \textit{which aspects of the input} are salient to particular model decisions.

Let a \textbf{cue} $C \subseteq x$ denote a contiguous span of an input $x$ hypothesized to causally influence a model's behavior.
We write $x_{\setminus C}$ for the \textbf{cue-ablated input} obtained by removing $C$ from $x$, and refer to the original $x$ as the \textbf{cue-included input}.
In general, we are interested in enabling LMs to answer questions of the form ``\textit{[$x$]. If the cue were removed, how would the assistant's answer change?}'' with answers of the form ``\textit{The response [would/would not] change to <$\M(x_{\setminus C})$>.}''
We train models to produce these answers %
via supervised fine-tuning (SFT). 
We use notation shown in~\Cref{tab:notation} throughout.
As illustrated in~\Cref{fig:overview}, our method has four steps:

\begin{enumerate}[leftmargin=*]
\setlength\itemsep{0.5pt}
    \item \textbf{\Mzero explanation construction (\S\ref{sec:methods:framework}):} We start with base model \Mzero, sample model behaviors $\BB{\Mzero}$, from which ground-truth explanations $\EE{\Mzero}$ are constructed.
    In the running example of~\Cref{fig:overview}, \Mzero answers A with the hint and B without it, so $\EE{\Mzero}$ is ``behavior changes from A to B when the hint is removed''.
    \item \textbf{Explanation training (\S\ref{sec:methods:training_recipe}):} We train \Mzero on explanations $\EE{\Mzero}$ while regularizing its behaviors towards \BB{\Mzero}, 
    obtaining a model:
    \begin{equation}
    \label{eq:reg_training}
    \Mreg = \Mtrain{\BB{\Mzero},\EE{\Mzero}}.
    \end{equation}
    \item \textbf{\Mreg explanation construction (\S\ref{sec:methods:framework}):} We roll out behaviors $\BB{\Mreg}$ from the new $\Mreg$ and construct ground-truth explanations $\EE{\Mreg}$ describing those behaviors.
    In~\Cref{fig:overview}, \Mreg answers B both with and without the hint, so $\EE{\Mreg}$ is ``behavior remains unchanged as B.''

    \item \textbf{Measuring introspective coupling (\S\ref{sec:methods:metrics}):} We evaluate whether the predicted explanations generated by \Mreg better match the ground-truth explanations of the original model $\EE{\Mzero}$, or the explanations of self $\EE{\Mreg}$.
    As shorthand, we denote evaluating \Mreg's predictions on \Mzero's explanations as $\Mreg \to \Mzero$ (or $\Orig$) and evaluating \Mreg's predictions on \Mreg's explanations as $\Mreg \to \Mreg$ (or $\Self$). \textbf{When \Self > \Orig, we say that a model exhibits introspective coupling.}
    In~\Cref{fig:overview}: $\Mreg$'s output matches $\EE{\Mreg}$ but not $\EE{\Mzero}$, and thus, introspective coupling is observed.
\end{enumerate}

\subsection{Explanation Training}
\label{sec:methods:training_recipe}
Given a model \M,
we perform SFT training on  explanation data from base model \EE{\Mzero} while regularizing behaviors to be close to \BB{\Mzero}. 
We use cross-entropy to train explanations and KL divergence to regularize behaviors.
Formally, the objective is:
\begin{equation}
\mathcal{L}(\mathcal{M}) =
\underbrace{\mathbb{E}_{(\metaq,\metae)\sim \EE{\Mzero}}\!\big[
-\log p_{\M}(\metae\mid \metaq)
\big]}_{\text{explanation cross-entropy}}
\,+\,
\lambda\,
\underbrace{\mathbb{E}_{(x,y)\sim \BB{\Mzero}}\!\big[
\mathrm{KL}(p_{\Mzero}(y\mid x) \,\|\, p_\M(y\mid x))
\big]}_{\text{behavioral regularizer}}
\label{eq:objective_decomp}
\end{equation}

\subsection{Evaluation Metrics}
\label{sec:methods:metrics}
We evaluate all models on two main metrics: \textbf{(1) Explanation Match (EM)} scores how well the LM \M has learned to explain a target model $\M'$, where $\M'=\Mzero$ for Orig and $\M'=\Mreg$ for Self. 
Specifically, we measure exact match between the explainer \M's prediction $\metaehat^{(\M)} \sim \M(\cdot \mid \metaq)$ for each meta-question $\metaq$ (``if the cue were removed, how would your answer change?'') and the ground-truth explanation in $\EE{\M'}$, averaged over $\EE{\M'}$.

\textbf{(2) Behavioral Match (BM)} captures the drift between two models' behavioral outputs, such as the drift of a model before and after explanation training. 
Let $\mathcal{D} = \{(x, C)\}$ be a held-out set of inputs $x$ and cue spans $C$, and let $y_\M(x) \sim \M(\cdot \mid x)$ denote $\M$'s label on $x$, and $y_\M(x_{\setminus C})$ denote the label on $x_{\setminus C}$.
We measure agreement between $\M$ and $\Mzero$ on both versions of each input:
\begin{equation}
\mathrm{BM}(\mathcal{M}, \mathcal{M}_0) \;=\; \frac{1}{|\mathcal{D}|}\sum_{(x,C)\in\mathcal{D}}
\underbrace{\mathbf{1}\!\left[y_{\M}(x) = y_{\Mzero}(x)\right]}_{\text{cue-included match}} \cdot
\underbrace{\mathbf{1}\!\left[y_{\M}(x_{\setminus C}) = y_{\Mzero}(x_{\setminus C})\right]}_{\text{cue-ablated match}}.
\end{equation}
In the metric above, both sides of the behavior must match. 
Fine-grained metrics are in~\Cref{sec:appendix:fine_grained_metrics}.
We report error bars as 95\% bootstrap CIs and significance with exact McNemar tests.

\section{Characterizing Introspective Coupling}
\label{sec:self_vs_original}

\subsection{Experiment Setup}
\label{sec:self_vs_original:setup}
\textbf{Models.} Our primary model $\Mzero$ is \textbf{Qwen3-8B}~\citep{yang2025qwen3technicalreport}.
To verify that our result is not model-specific, we show results across model families and sizes in~\Cref{sec:behavior_distribution:llama_hint}.
We use full fine-tuning by default and discuss LoRA fine-tuning in~\Cref{sec:appendix:lora}. 

\textbf{Sycophancy Datasets.}
\label{sec:methods:tasks}
Modern language models often exhibit sycophancy---over-agreeableness with the user---%
at the expense of accuracy~\citep{sharma2025understandingsycophancylanguagemodels,wei2024simplesyntheticdatareduces}.
We investigate training models to articulate \textit{when} they are sycophantic.
Specifically, we study two datasets:
First, \textbf{Hint-MMLU}~\citep{chen2025reasoningmodelsdontsay,li2025traininglanguagemodelsexplain} assesses whether models will modify their answer to follow a user-suggested hint.
Here, $x$ is an MMLU multiple-choice question with an injected $C=$``Hint: \texttt{A}'' string (see~\Cref{fig:overview}).
Sycophantic models will change their answer when the hint is present; we train models to articulate whether they will do so.
Second, \textbf{AITA} measures whether models are overly inclined to flatter the user~\citep{bsher2023aitageneratingmoraljudgements}. This dataset consists of morally ambiguous stories from Reddit's \textit{r/AmItheAsshole}. 
Here, $x$ is a Reddit post and $y$ is a judgment of the user's moral character $\in \{\text{``Not the Asshole (NTA)''},
\text{``You're the
Asshole (YTA)''}\}$. 
$C$ is a system prompt %
framing the story in first- or third-person. 
In this case, sycophantic models will change their judgment to be more favorable if the story were written in first-person. We train models to articulate when they will change their judgment based on the system prompt.

\paragraph{Refusal Datasets.}  
Understanding when LMs refuse requests allows us to debug %
over- or under-refusals.
We specifically investigate how the user's presented \textit{role} interacts with refusal. The user's role is a popular attack surface---models may be more susceptible to answering harmful queries for users who are fiction writers, or to overrefusing users who are kindergarten teachers.
Thus, we would like to train models to articulate whether and how their refusal hinges on a stated role.
$x$ is a user prompt and $C$ is a system prompt that frames the user in different roles. See examples in~\Cref{sec:behavior_distribution:refusal}.
We measure $y \in \{\text{refuse}, \text{comply}\}$ by
sampling a freeform response %
and scoring it with an LLM
judge. 
Note that this setting is much harder than Hint-MMLU and AITA: the behavior label is inferred from freeform generation rather than constructed deterministically from a single
answer token, so the explanation target is a higher-level property of the response.
We obtain $x$ by combining two refusal datasets: \textbf{FalseReject}~\citep{zhang2025falsereject}, a benign dataset to test over-refusal, and
\textbf{WildJailbreak}~\citep{jiang2024wildteaming} for harmful queries. We generate roleplay system prompts $C$ with an LLM.

\subsection{Results}
\label{sec:self_vs_original:results}

\begin{figure*}
    \centering
    \includegraphics[width=\linewidth]{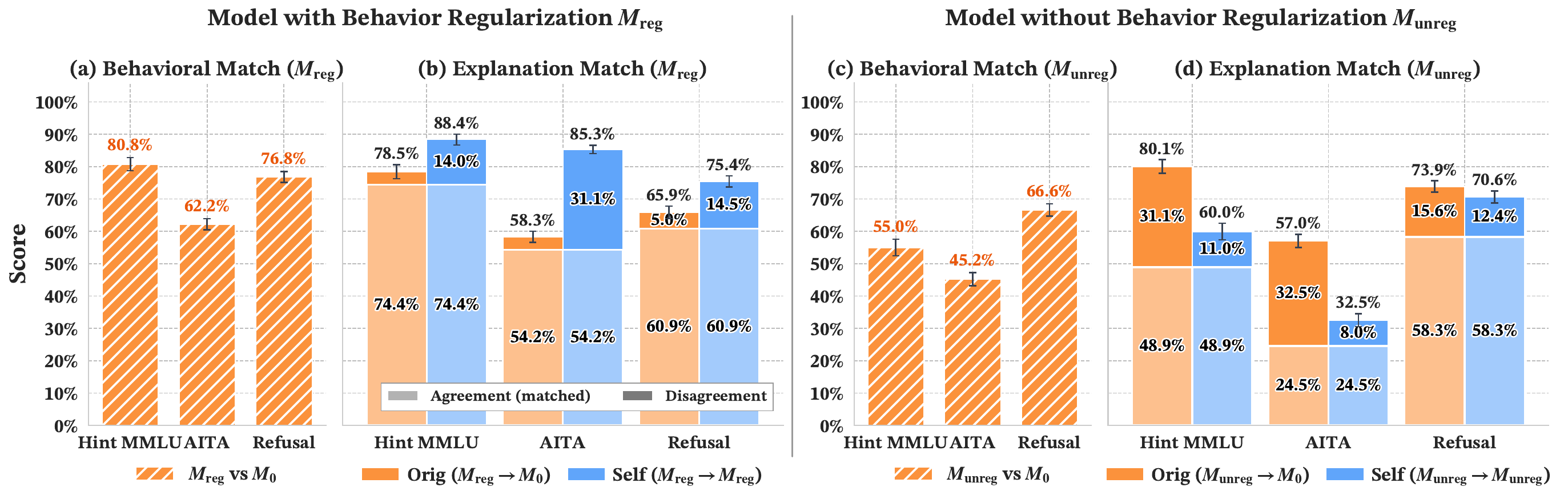}
    \caption{
    \Self~$>$~\Orig emerges only with regularization, across Hint-MMLU, AITA, and Refusal domains.
    (\textbf{a}, \textbf{b}): the regularized model \Mreg.
    (\textbf{c}, \textbf{d}): the unregularized model \Munreg.
    \textbf{(a, c)} Behavioral Match: agreement between original labels
    $\BB{\Mzero}$ and labels after training ($\BB{\Mreg}$ and $\BB{\Munreg}$ respectively).
    \textbf{(b, d)} Explanation Match: explanations against
    $\EE{\Mzero}$ labels (orange) and self labels (blue). \Mreg, despite being trained on \Mzero, explains itself better than \Mzero (b);
    without regularization, this gap reverses (d).
    Much of $\Mreg$'s apparent ability to explain \Mzero comes from the subset where \Mzero and \Mreg \textit{agree} (the lighter lower segment in panel b), and the gap comes from the disagreement subset.
    }    
    \label{fig:self_vs_orig}
    \vspace{-1em}
\end{figure*}

Results can be found in~\Cref{fig:self_vs_orig}, where we compare regularized training ($\lambda=1$ in~\Cref{eq:objective_decomp}, left side) against unregularized training ($\lambda=0$ in~\Cref{eq:objective_decomp}, right side), plotting both behavior drift (a,c) and Explanation Match for both Self and Orig (b,d).

From \Cref{fig:self_vs_orig}(b), we observe that across held-out data on all three tasks---Hint-MMLU, AITA, and Refusal---the regularized model $\Mreg$'s predicted explanations match its own labels $\EE{\Mreg}$ better than the original
targets $\EE{\Mzero}$ it was trained on. Thus, we observe the
\textbf{\Self~$>$~\Orig} signature of introspective coupling 
across all three tasks.
By construction, this gap arises entirely from examples where the two labels disagree. This subset is plotted in \Cref{fig:self_vs_orig}(b) as a darker shade, above the white line. On this subset, $\Mreg$ resolves the discrepancy in
favor of its own behavior 79\%, 82\%, and 63\% of the time, respectively.
In \Cref{sec:appendix:not_heuristic}, we rule out trivial accounts explaining \textbf{\Self~$>$~\Orig}: that $\EE{\Mreg}$ drifts to a degenerate
distribution that is easier for \emph{any} external explainer to learn. We validate that regularized training produces this signature across model families and sizes (\Cref{fig:appendix_llama_hint_metrics}).

From \Cref{fig:self_vs_orig}(d), we observe that \textbf{\Self~$>$~\Orig} disappears without regularization:
a model trained on explanations
alone, $\Munreg = \M[\EE{\Mzero}]$, actually demonstrates Self < Orig.
When comparing \Cref{fig:self_vs_orig}(a) against \Cref{fig:self_vs_orig}(c), we notice that \Munreg's behavior drifts further away from \Mzero than \Mreg's behavior drifts.
Following this, we hypothesize that
behavior similarity during training governs coupling, which we
test in depth in \Cref{sec:when}.

\subsection{Introspective coupling at the mechanistic level}
\label{sec:self_vs_original:interpretability}

One hypothesis for why coupling emerges is that through training, computation that \emph{reports} the model's behavior learns to reuse the computation that \emph{produces} the behavior. 
We investigate whether the two representations are causally linked~\citep{merullo2024circuitcomponentreusetasks} at the mechanistic level, by
testing if
intervening on representations to shift behavior $B(\M)$ also shifts the explanation $E(\M)$ in the same direction.

\begin{figure*}[t]
    \centering
    \includegraphics[width=0.8\linewidth]{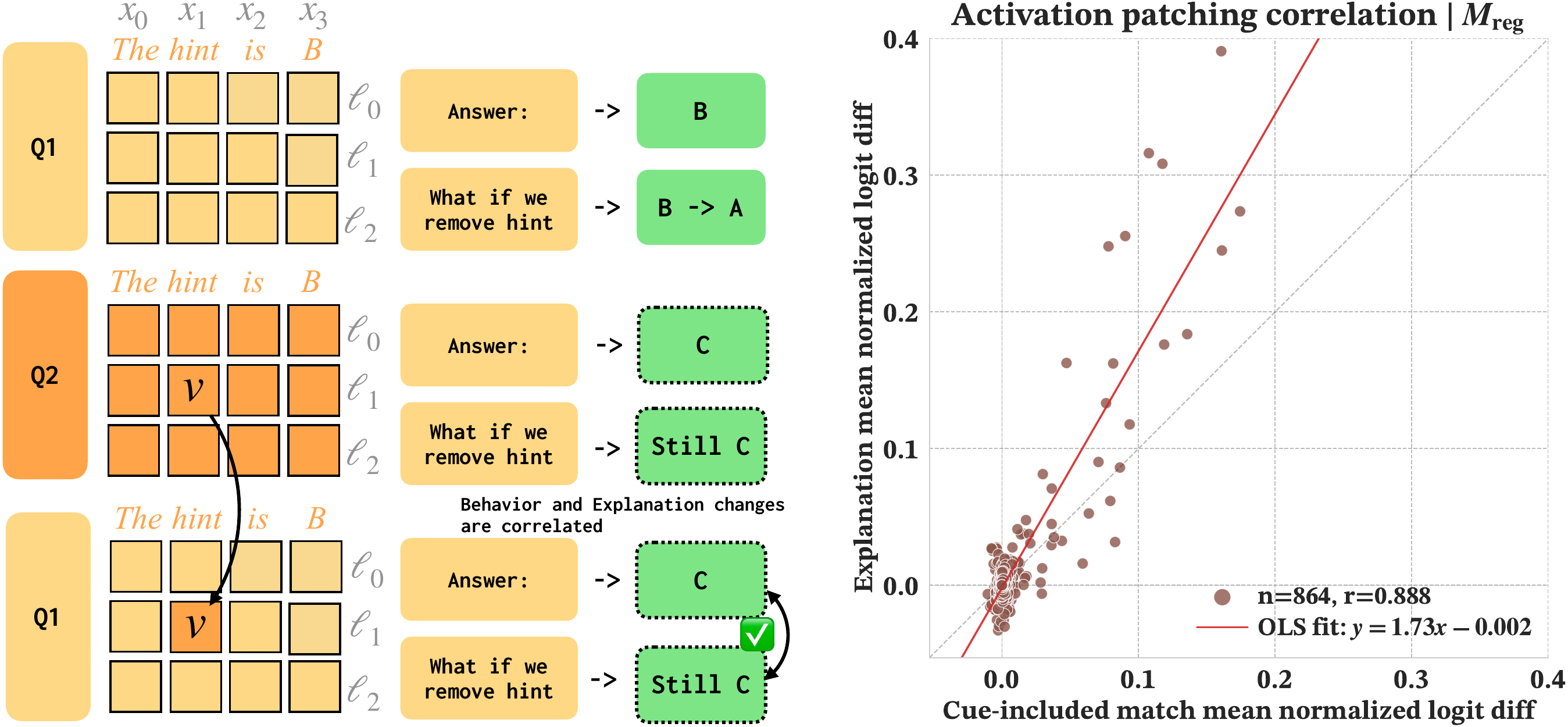}
    \caption{
    Mechanistic signature of introspection: activation-level interventions that change behavior also change the explanation.
    \textbf{Left:} we construct
      prompt pairs with the same cue but different explanation
    labels (removing the cue changes the answer for one, but not the other). We patch activations between them and measure whether
    shifts in behavior logits correlate with shifts in explanation logits.
    \textbf{Right:} Each point is one (layer, token) patching position. X-axis is the mean normalized logit difference for behavior
    (cue-included answer), while the y-axis shows the corresponding
    logit-difference for the explanation's change/no-change prediction. 
    Patching effects on
    behavior and explanation are strongly correlated with Pearson $r=0.89$.}
    \label{fig:activation_patching}
    \vspace{-1em}
\end{figure*}

We test this on Hint-MMLU models via activation patching (\Cref{fig:activation_patching} left). 
Recall that each explanation query $\metaq$ implicitly contains the underlying object-level input $x$ (\S\ref{sec:methods:framework}).
We patch activations into the shared prefix $x$ and measure the downstream effect on both behaviors $\M(x)$ and explanations  $\M(\metaq)$. 
We create two counterfactual inputs, a source $x$ and a target $x'$ (with corresponding meta-questions $\metaq,\metaq'$) that share the same hint cue $C=C'$, but 
differ in behavior, $\M(x)\neq\M(x')$, and hence in explanation, $\EE{\M}(\metaq)\neq\EE{\M}(\metaq')$.
We patch activations from $x$ to $x'$ (and from $\metaq$ into $\metaq'$) at corresponding
(layer, token) positions, %
and 
measure the normalized logit difference (NLD)~\citep{wang2023interpretability}
\begin{equation}
\mathrm{NLD}(u) = \frac{\ell_{\text{patched}} - \ell_{u'}}{\ell_{u} - \ell_{u'}}, \qquad u \in \{x, \metaq\},
\label{eq:nld}
\end{equation}
where $\ell$ is the output logit of each forward pass.
The NLD is 0 if the patch has no effect and 1 if it fully recovers the source output.

We validate the following hypothesis: if the model shares circuits between behavior and explanation, then patching interventions that affect behavioral logits $\mathrm{NLD}(x)$ are correlated with ones that affect explanation logits $\mathrm{NLD}(\metaq)$.
The result is shown in~\Cref{fig:activation_patching}, where interventions move the behavior logits and explanation logits in the same direction, with Pearson correlation $r=0.89$. This means that the same layer/token activations are used for behavioral responses as for the explanations.
Critically, the $\Munreg$ baseline, which has non-trivial Orig Explanation Match (\Cref{fig:self_vs_orig}(d)) but does not exhibit \textbf{\Self > \Orig}, only exhibits a correlation of $r=0.527$ (\Cref{sec:appendix:activation_patching}).

\section{When does introspective coupling happen?}
\label{sec:when}

Introspective coupling is desirable for faithful explanation because it means that the model's explanations track its current behavior. We next ask \textbf{when this coupling emerges.} Our findings in~\Cref{sec:self_vs_original:results} suggested that models only demonstrated Self > Orig under regularization, which reduced behavioral drift compared to no regularization.
Thus, we hypothesize that
coupling depends on whether the explanation supervision remains \textbf{behaviorally compatible} with the model throughout training.

Formally, let $E_{\mathrm{sup}}^{(t)}$ denote the explanation supervision labels used for SFT at training step $t$. In the default fixed-label setting, $E_{\mathrm{sup}}^{(t)} = \EE{\Mzero}$ for all $t$; in this section, we vary $E_{\mathrm{sup}}^{(t)}$ directly. Let $\EE{\M_t}$ denote the ground-truth explanations constructed from the current model's behavior at step $t$. We refer to the similarity between $E_{\mathrm{sup}}^{(t)}$ and $\EE{\M_t}$ as \textbf{online label--self agreement}: the degree to which
the explanation training targets at step $t$ agree with the explanations of the current model's behavior. Thus, our central hypothesis is
\begin{claim}
\textbf{Hypothesis:} High %
online label-self agreement (i.e. $E_{\mathrm{sup}}^{(t)}$ and $\EE{\M_t}$ agreement) over the course of training (i.e. $\forall t$) is an important factor for the emergence of $\Self > \Orig$. \end{claim}
The three subsections below probe this hypothesis by varying online agreement through different mechanisms: changing the behavioral regularization weight $\lambda$, which affects how close $\M_t$ remains to the fixed explanation labels (\Cref{sec:when:reg_sweep}); directly controlling the agreement between $E_{\mathrm{sup}}^{(t)}$ and $\EE{\M_t}$ via online relabeling (\Cref{sec:when:continuous_relabel}); and training on explanation targets from a different initial model (\Cref{sec:when:Llama_mix}), which separates the role of \emph{initial} label--model similarity from \emph{online} compatibility during training.\footnote{We also plot the emergence of coupling over the course of training in \Cref{sec:when:train_dynamics}.} 
Finally, we also perform a learning rate sweep that uncovers evidence \textit{against} this hypothesis, suggesting that multiple factors may be necessary to induce Self > Orig (\Cref{sec:appendix:hint_lr_sweep}).
We focus our analysis in this section on the Hint-MMLU setting.

\subsection{Behavioral Regularization Preserves Online Label--Self Agreement}
\label{sec:when:reg_sweep}

\begin{figure*}[t]
    \centering
    \begin{minipage}[t]{0.49\linewidth}
        \centering
        \includegraphics[width=\linewidth]{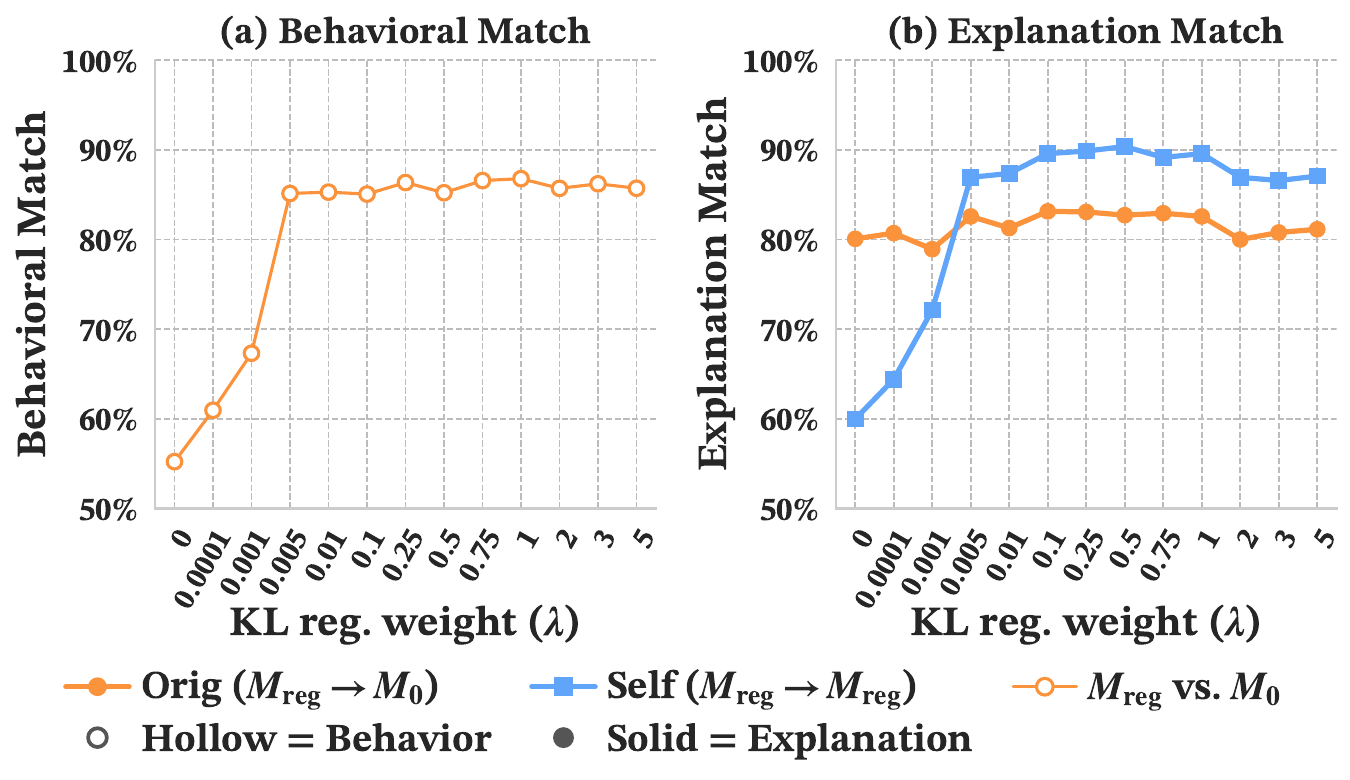}
        \vspace{-1.5em}
        \caption{ Behavioral regularization weight $\lambda$ sweep on Hint-MMLU (\S\ref{sec:when:reg_sweep}). \textbf{(a)} Behavioral Match between \Mreg and \Mzero rises at the same $\lambda$-values as \Self Explanation Match shown in \textbf{(b)}, which shows the  \Self > \Orig signature and overtakes \Orig once $\lambda\ge 5\mathrm{e}{-3}$ and stays ahead for bigger $\lambda$.}
        \label{fig:reg_sweep}
    \end{minipage}\hfill
    \begin{minipage}[t]{0.49\linewidth}
        \centering
        \includegraphics[width=\linewidth]{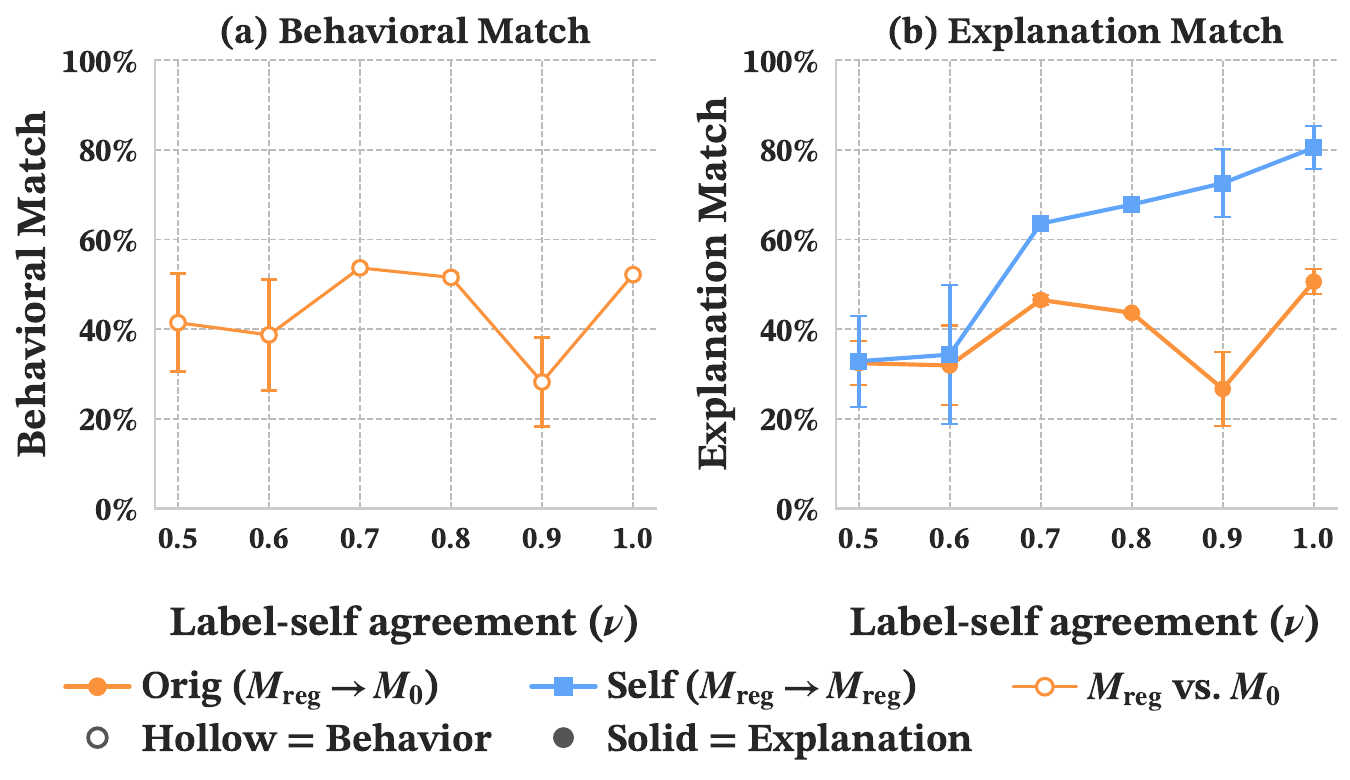}
        \vspace{-1.5em}
        \caption{Continuous relabeling with  adjustable label--self agreement $\nu$ between training targets $E_{\mathrm{sup}}^{(t)}$ and self-labels $\EE{\M_t}$ (\S\ref{sec:when:continuous_relabel}). \textbf{(a)} Behavioral Match stays constant at $\approx 0.5$, so \Self $>$ \Orig does not come from reduced drift from $\Mzero$. \textbf{(b)} Meanwhile, Self Explanation Match rises sharply over \Orig once $\nu \approx 0.7$.}
        \label{fig:continuous_relabel}
    \end{minipage}
    \vspace{-0.5em}
\end{figure*}

We begin by performing a finer-grained version of our regularization vs. no regularization experiment in~\Cref{sec:self_vs_original:results}, by varying the \textit{weight} $\lambda$ on the behavioral regularization term in our objective (\Cref{eq:objective_decomp}).
We sweep $\lambda$ across five orders of magnitude, 
with results shown in \Cref{fig:reg_sweep}. We find that coupling begins to emerge at remarkably small values: a \Self~$>$~\Orig gap is visible for any $\lambda \gtrsim 5\!\times\!10^{-3}$, and the gap stays consistent for bigger $\lambda$.
Notably, this inflection coincides with a large increase in Behavioral Match
(\Cref{fig:reg_sweep}a) between $\M_t$ and $\Mzero$.
This is consistent with our hypothesis that high online label-self agreement is important for Self > Orig. 
 In this setup, the original model $\Mzero$ is the supervision source throughout training so $E_\text{sup}^{(t)} = \EE{\Mzero}$ for all $t$. 
Thus, when $\M_t$ and $\Mzero$ remain close (high Behavioral Match) throughout training, $\EE{\M_t}$ remains similar to $E_\text{sup}^{(t)}$, and Self > Orig emerges. 
However, this experiment does not distinguish whether coupling requires the current model being trained to remain close to the original model $\Mzero$ or to the source of the explanation labels $E_\text{sup}^{(t)}$. We isolate this distinction directly in \Cref{sec:when:continuous_relabel}, where we vary the current model's proximity to the labels, independent of the original model.

\subsection{Introspective Coupling Depends on Online Label-Self Agreement}
\label{sec:when:continuous_relabel}

We conduct a controlled test of our hypothesis
that high online label-self agreement leads to introspective coupling.
Because $\M$'s behavior changes dramatically over the course of training, 
at every gradient step, we regenerate the explanation training targets $E_{\mathrm{sup}}^{(t)}$ so that they agree with the \textit{current} model behavior $\BB{\M}$ on a fraction $\nu$ of labels and disagree---via a randomly sampled alternative answer---on the remaining fraction $1-\nu$:
$$E_{\mathrm{sup}}^{(t)}
=
\left\{
(\metaq, \metae) \;\middle|\;
\metae =
\begin{cases}
\BB{\M}(\metaq) & \text{with probability } \nu \\
\tilde{\metae} \neq \BB{\M}(\metaq) & \text{with probability } 1 - \nu
\end{cases}
\right\}$$
On the behavioral side, we still regularize $\M$ toward the base model $\Mzero$. Thus, we have a model:
\[
\M_{\textrm{noise-self}} = \Mtrain{E_{\mathrm{sup}}^{(t)},\BB\Mzero}.
\]
We sweep $\nu$ in
\Cref{fig:continuous_relabel}.
In (b), we find a sharp transition in explanation behavior as the online
label--self agreement $\nu$ varies. 
We observe introspective coupling (\Self > \Orig) for $\nu\ge 0.7$ but 
once $\nu \le 0.6$, both \Self and \Orig Explanation Match collapse. 
This supports
our hypothesis that coupling requires a minimum level of agreement between the
explanation supervision and the current model's behavior; for Qwen3-8B on
Hint-MMLU, this threshold also appears to be $\approx 0.7$.

We measure Behavioral Match between the trained model and
the original model \Mzero in~\Cref{fig:continuous_relabel}(a). Interestingly, Behavioral Match remains roughly
constant across $\nu$. 
This disentangles the two factors we suggested at the end of the previous section: staying close to the \textit{explanation labels} during training matters more than staying close to the \textit{original model}.

\subsection{Introspective Coupling Persists with Cross-Model Explanation Labels}
\label{sec:when:Llama_mix}
\begin{figure*}[t]
    \centering
    \includegraphics[width=\linewidth]{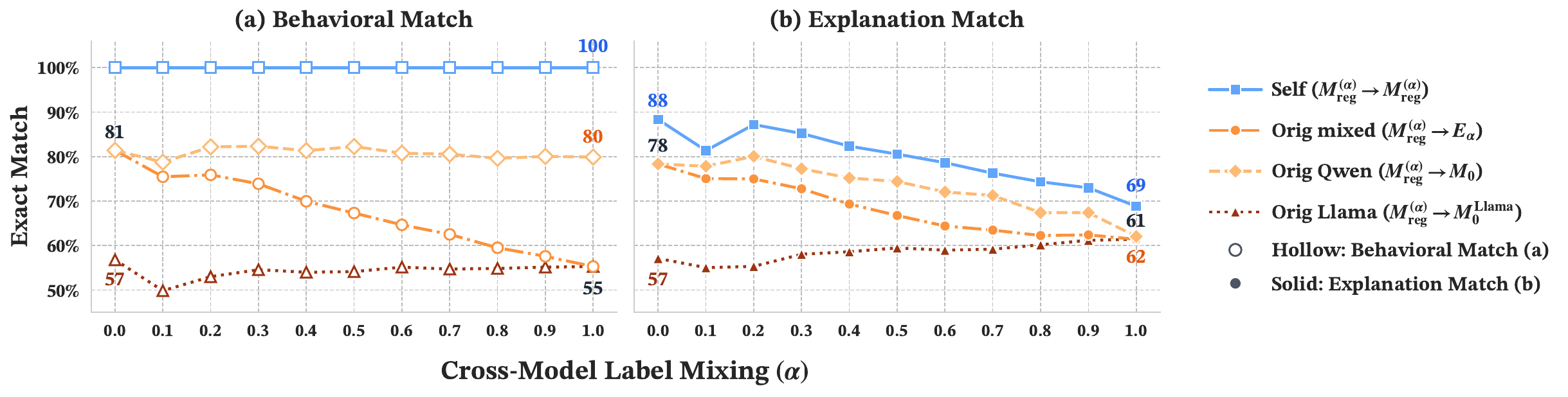}
    \vspace{-1em}
    \caption{
    Training on explanation labels from another model (\S\ref{sec:when:Llama_mix}).
    Explanation labels $E_\alpha$ mix a fraction of Llama-3.1-8B-Instruct labels into Qwen3-8B's labels ($\alpha = 0$: all Qwen's labels, $\alpha=1$: all Llama's). Behavior
    is still regularized toward Qwen.
    We score each trained model
    against four label sources: \Self, $E_\alpha$ (mixed), Qwen-base, and
    Llama-base.
    \textbf{(a)} Across the sweep, behavior remains close to initial Qwen ($\approx$80\% Behavioral Match), and \textbf{(b)}  \Self > \Orig holds at every $\alpha$. Though explanation
    supervision comes from another model, the trained model's
    explanations continue to track its own current behavior better than either
    base-model label source.
     }
    \label{fig:Llama_mix_alpha_sweep}
\end{figure*}

Previously, we found that it is more important for the current model to remain close to the explanation labels than to the initial model.
We next ask whether those labels must be generated by the model's own initial checkpoint. \textbf{Can a model learn to self-explain from labels generated by a different model, without substantial behavioral drift?}
If so, explanation labels need not be
collected separately for every model: labels from one model could induce
self-explanations in another.

To test this, we start with a Qwen3-8B base model \Mzero, from which we derive
behaviors $\BB{\Mzero}$ and ground-truth explanations $\EE{\Mzero}$. 
We then replace some of Qwen's explanation labels with labels from Llama-3.1-8B-Instruct, denoted \MLlama.
The two models' ground-truth explanations agree on only 53\% of examples.

For each mixture ratio $\alpha \in [0,1]$, we construct an explanation supervision dataset
\[
E_\alpha = (1-\alpha)\,\EE{\Mzero}
+ \alpha\,\EE{\MLlama},
\]
where $\alpha$ is the fraction of explanation labels sourced from Llama. We train
\[
\Mmix \;=\; \M\!\left[\BB{\Mzero},\; E_\alpha\right].
\]
Crucially, we
regularize behavior toward the original Qwen throughout, using $\BB{\Mzero}$; only the
explanation supervision is mixed. Thus, $\alpha=0$ corresponds to pure Qwen
explanation supervision, while $\alpha=1$ corresponds to pure Llama explanation
supervision with Qwen behavioral regularization.

Results are shown in~\Cref{fig:Llama_mix_alpha_sweep}. 
Panel (a) confirms that this intervention does not significantly affect the trained model's behavior. Across $\alpha$, $\Mreg^{(\alpha)}$ remains a steady 80\% from \Mzero (base Qwen) and a steady 55\% from \MLlama (base Llama). Thus, replacing Qwen explanation labels with Llama labels does not cause the trained model's behavior to move toward Llama.
Nevertheless, on the explanation side (b), \Self~$>$~\Orig holds
for every value of $\alpha$. Thus, \textbf{introspective coupling does not require explanation labels to come from the same model being trained}.\footnote{That said, explanation quality does degrade overall as $\alpha$ increases: \Self Explanation Match falls from $88\%$ ($\alpha=0.0$) to
$69\%$ ($\alpha=1.0$), as expected if Llama labels act as noise to the function of modeling Qwen's own behaviors.
}

These results suggest that explanation datasets may be reusable across
behaviorally similar models. 
Further, these results add to a growing body of work arguing that self-explanation in LMs leverages \emph{privileged access}
\citep{binder2025looking,li2025traininglanguagemodelsexplain}. Even when the
model is explicitly trained on another model's explanations, including examples
where the two models disagree, its predictions track its own behavior more
closely than the foreign explanation labels. We interpret this as 
evidence that introspective coupling depends more on \textit{online label--self
agreement} than on the source of the original explanation labels.

\section{Generalization}
\label{sec:generalization}
\newcommand{\Maux}{\ensuremath{\mathcal{M}_\mathrm{aux}}\xspace}
In~\Cref{sec:self_vs_original,sec:when}, we studied behavioral drift that occurs as an emergent consequence of explanation training.
However, in realistic post-training scenarios (or in scenarios where explanations would be useful), %
models are trained concurrently with other post-training data that may induce different behavioral shifts.
\textbf{We study how well introspective coupling generalizes to
explaining new or shifted behaviors that the model acquired through concurrent post-training.}\footnote{A version of this question was explored by~\citet{binder2025looking} using synthetic tasks; we focus on more complex explanations with direct safety relevance, such as sycophancy or refusal behavior, under realistic post-training mixtures.}
To formalize this setup, we extend our notation to include a dataset argument: let $\BB{\M,\mathcal{D}}$ denote behaviors produced by model $\M$ on dataset $\mathcal{D}$ (e.g. Hint-MMLU), and let $\EE{\M,\mathcal{D}}$ denote explanations corresponding to those behaviors.
Starting from our default training pipeline, we concurrently train \M on an auxiliary post-training dataset $A$:
\begin{equation}
\label{eq:aux_training}
\Maux \;=\; \M\!\left[\BB{\Mzero, \mathcal{D}},\; \EE{\Mzero, \mathcal{D}}, \; A\right]
\end{equation}
where explanation training and behavioral regularization are applied only on the original dataset $\mathcal{D}$ and the auxiliary dataset $A$ provides \emph{behavior-only} supervision, not explanation supervision.

We then evaluate whether $\Maux$'s explanations track its current behavior on $\mathcal{D}$ and on $A$. Ground-truth explanations are constructed post hoc from $\Maux$'s own behavior. Thus, faithful explanations on $A$ cannot result from direct supervision and must reflect generalization of the explanation function.
We study two kinds of auxiliary shift: in \S\ref{sec:generalization:jabberwocky} we inject new, synthetic behavior, and in \S\ref{sec:generalization:behavior_drift} we design a realistic post-training corpus that shifts existing behaviors.

\subsection{Generalization to Newly Acquired Behaviors}
\label{sec:generalization:jabberwocky}
\begin{figure*}[t]
    \centering
    \vspace{-0.8em}
    \includegraphics[width=\linewidth]{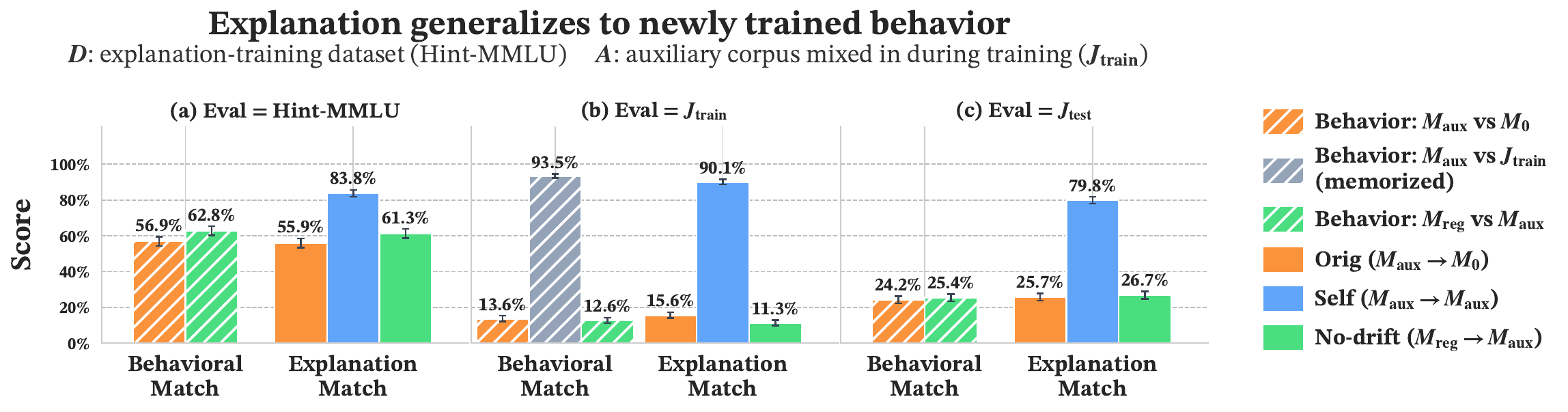}
    \caption{
    Explanations generalize to newly acquired behaviors (\S\ref{sec:generalization:jabberwocky}). We train $\Maux$ with explanation supervision on Hint-MMLU ($D$) plus behavior-only training on gibberish data \Jtrain{} ($A$), and evaluate on held-out Hint-MMLU, \Jtrain{}, and \Jtest{}. \Self~(blue)~>~\Orig~(orange) holds on all three sets.
    Although it never saw \Jtrain{} \emph{explanations} during training, nor \Jtest{} explanations or behavior, $\Maux$ explains both at high accuracy. $\Mreg$, which never saw Jabberwocky, can explain $\Maux$ only at chance.
     }
    \label{fig:jabberwocky_tracking}
    \vspace{-0.75em}
\end{figure*}

\begin{figure*}
    \centering
    \includegraphics[width=\linewidth,trim={0 15pt 0 0},clip]{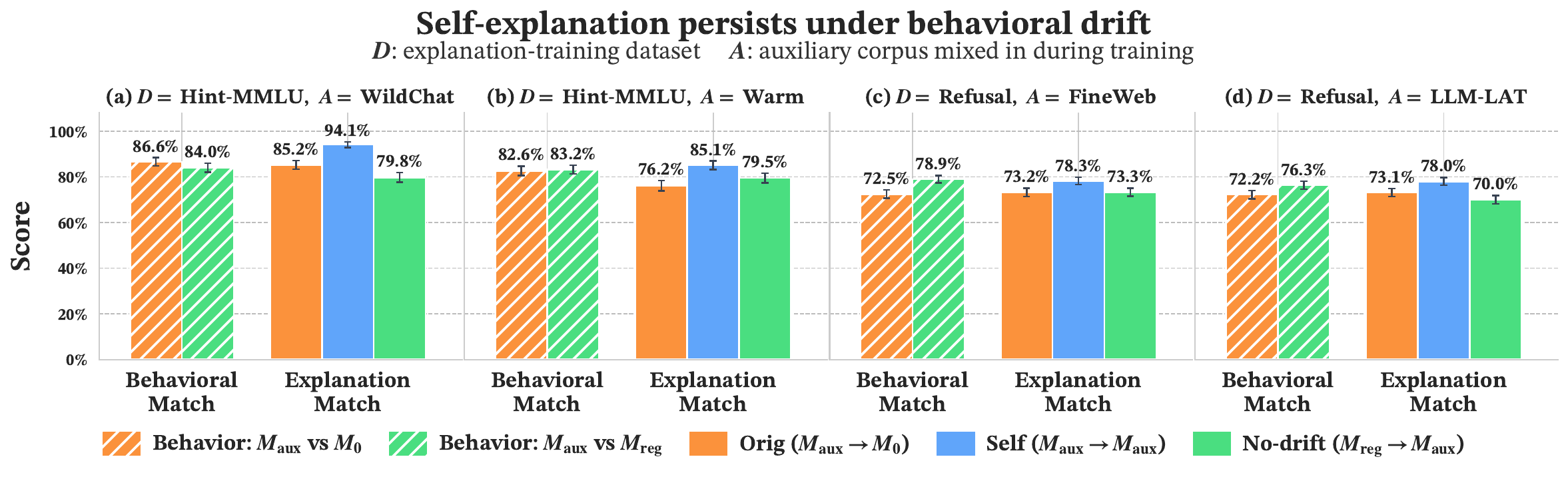}
    \caption{
    Explanations track shifts to existing behaviors (\S\ref{sec:generalization:behavior_drift}). We mix auxiliary post-training corpora into explanation training (\Cref{eq:aux_training}): each panel \textbf{(a)--(d)} mixes an explanation-training dataset $\mathcal{D}$ with an auxiliary corpus $A$. \textbf{Behavioral}: every setup drifts from both $\Mzero$ and $\Mreg$ (explanation training only). \textbf{Explanation}: We find that (1) \Self > \Orig persists under auxiliary training, and (2) \Self > $\Mreg\to\Maux$, showing that the model's explanations have shifted alongside auxiliary training to reflect its shifted behavior.
     }
    \label{fig:combined_self_explainer}
\end{figure*}

We first test whether coupling extends to behaviors the model could only have acquired through auxiliary training.
We generate synthetic nonsense ``Jabberwocky'' data $J = \{(x_J,y_J)\}$ that follows the Hint-MMLU question format but uses
fictitious, alien-sounding questions that LMs do not have priors over. The model's behavior on $J$ must be trained in (samples in~\Cref{sec:appendix:jabberwocky}).
We use the objective in~\Cref{eq:aux_training}, where we train on explanations from $D=$ Hint-MMLU and on the Jabberwocky auxiliary dataset  $A=\Jtrain$, holding out $\Jtest$ to test generalization. 

\paragraph{Results.}
\Cref{fig:jabberwocky_tracking} reports Explanation Match on the three evaluation sets ($\mathcal{D},A,\Jtest$).
First we observe \Self (blue) > \Orig (orange) on all three datasets, \textbf{indicating that introspective coupling generalizes to auxiliary behavioral signal}.
In particular, (a) shows that the coupling on the original dataset $\mathcal{D}$ is preserved under auxiliary signal, (b) shows that $\Maux$ explains \Jtrain nearly perfectly (90.5\%) despite never receiving explanation supervision, and (c) shows that explanation accuracy is 79.8\% on \Jtest, which received neither explanation nor behavior signal.\footnote{\Jtest responses are near random and have no ground-truth, and the model defers to changing to the hint. However, \Maux knows to output the \textit{cue-ablated} random response, an ability that \Mzero or \Mreg don't have. Details in~\Cref{sec:appendix:jabberwocky}.}
These results are not explained by the auxiliary data failing to change the model: $\Mreg$, which never sees Jabberwocky data, explains $\Maux$ only at chance. Thus, $\Maux$'s explanations track behaviors acquired through additional training rather than merely preserving its original explanation policy.

\subsection{Explanations Track Shifts to Existing Behaviors}
\label{sec:generalization:behavior_drift}
While~\Cref{sec:generalization:jabberwocky} tested whether explanations extend to a \emph{new} behavioral domain, we now test whether they track changes that auxiliary training induces \emph{within} the original explanation domain $\mathcal{D}$. %
Post-training routinely shifts behavior in unintentional ways: instruction-tuning increases sycophancy~\citep{wei2024simplesyntheticdatareduces}, and narrow fine-tuning can produce broadly misaligned models that stop refusing~\citep{betley2025emergent}. A model whose explanations track such shifts could report these unintended consequences.
We reuse objective~\Cref{eq:aux_training}, now instantiating $A$ with real post-training corpora (details in \Cref{sec:appendix:drift_auxiliary_data}):

\begin{itemize}[leftmargin=*]
\setlength\itemsep{1pt}
    \item \textbf{$\mathcal{D}=$ Hint-MMLU, $A=$ WildChat~\citep{zhao2024wildchat}:} real user-LM interactions used for post-training.
    \item \textbf{$\mathcal{D}=$ Hint-MMLU, $A=$ Warm Responses~\citep{ibrahimwarm2026}:} empathetic dialogue that raises sycophancy.
    \item \textbf{$\mathcal{D}=$ Refusal, $A=$ FineWeb~\citep{penedo2024the}:} a generic web corpus inducing a less directed behavioral shift.
    \item \textbf{$\mathcal{D}=$ Refusal, $A=$ LLM-LAT Direct Refusal~\citep{sheshadri2025latent}:} harmful requests paired with direct refusals.
\end{itemize}

\Cref{fig:combined_self_explainer} confirms that all four setups produce meaningful drift despite behavioral regularization, with respect to both the base model $\Mzero$, and a model $\Mreg$ without training on auxiliary data $A$.
Furthermore, 
we find that: %
(1) \textbf{\Self~$>$~\Orig persists under auxiliary training}: $\Maux\to\Maux$ (blue) beats $\Maux\to\Mzero$ (orange) across all settings, and 
(2) \textbf{Explanations track with auxiliary-training-induced shift}:
$\Maux$ explains its own behavior better than the no-drift explainer $\Mreg$ does, indicating that $\Maux$ is not simply learning to 
explain the non-drifted part of its behavior.

These results have a practical implication for monitoring unintended consequences of post-training: when explanation training runs alongside other post-training, a model can report on the behavioral shifts it acquires during training.
Self-explanation could help cheaply monitor emergent behavioral change throughout post-training without new explanation labels.

\section{Related Work}
\label{sec:related_work}
LMs can produce natural language explanations of their outputs, either in chain-of-thought (CoT) or post-hoc, but their explanations can be unfaithful to %
the true decision-making processes
~\citep{turpin2023language,lanham2023measuringfaithfulnesschainofthoughtreasoning,barez2025cot}. 
Faithful CoTs are useful for external monitoring in safety-critical scenarios~\citep{korbak2025chainthoughtmonitorabilitynew,guan2025monitoringmonitorability}. This objective can be thought of as a self-consistency objective between model verbalizations and behaviors~\citep{pres2026position}. Prior work has studied explanations of model behaviors~\citep{joglekar2025trainingllmshonestyconfessions,li2026spillingbeansteachingllms,hase2026counterfactual,mayne2026positive}, internal activations~\citep{pan2024latentqa,karvonen2026activationoraclestrainingevaluating,frasertaliente2026nla,huang2025predictive,choi2025scalably}, and training data~\citep{goel2025learninginterpretweightdifferences,shenoy2026introspectionadapterstrainingllms}.
Closest to our work, \citet{hase2026counterfactual} and \citet{turpin2025verbalization} train CoT to verbalize hint-style cues, whereas we train meta-level predictions of counterfactual behavior without CoT.

Recent work also investigates whether models possess \textit{metacognition} or \textit{introspective abilities}, either zero-shot~\citep{comsa2025doesmakesensespeak,laine2024memyselfaisituational,lindsey2025emergent,zhong2026spontaneous} or through fine-tuning~\citep{binder2025looking,plunkett2025selfinterpretabilityllmscomplexinternal,betley2025tellyourselfllmsaware}. One set of work focuses on whether the model has self-knowledge of external tampering~\citep{lindsey2025emergent,macar2026mechanismsintrospectiveawareness,pearsonvogel2026latentintrospectionmodelsdetect,lederman2026emergentintrospectionaicontentagnostic}, while another set investigates whether models can model their own output distribution~\citep{binder2025looking,li2025traininglanguagemodelsexplain,song2025language}. Our work falls into the second camp. 
A central question in the literature is whether models have privileged access when introspecting, or whether such introspective behaviors can be simulated by external models~\citep{song2025privilegedselfaccessmattersintrospection,li2025do,singh2026llmsintrospectrealitycheck}. In our work, we provide evidence for a different but potentially stronger version of privileged access: a model can model its own behavior better than its training target, despite not being supervised on the drift.

\section{Conclusion}

We have shown the surprising phenomenon of \emph{introspective coupling}: LMs trained on a fixed set of explanations derived from their base model learn to explain their \emph{own} current behaviors more faithfully than the training targets they were supervised on. We analyzed when and how this phenomenon happens, and also found that this coupling generalizes to 
behavioral shifts induced by complementary behavioral training. 
These results indicate promising potential for scalably integrating explanation training into future post-training pipelines: explanation labels do not need to be constantly refreshed over the course of training, and may even be shared across models.
Our setting is deliberately narrow: each model is trained and evaluated on a single cue-based task; whether broader explanation training yields similar generalization remains an open question.
We discuss detailed limitations and future work in~\Cref{sec:appendix:limitation} and broader impact in~\Cref{sec:appendix:broader_impact}.

\subsection*{Acknowledgments}
This work was supported by the National Science Foundation through grant
IIS-2238240, the IARPA BENGAL program, the DARPA AIQ program through CMO
contract HR00112520025, and the MIT Generative AI Consortium.
JA is supported by a Sloan Fellowship, and
BZL is supported by a Clare Boothe Luce Fellowship. 
We thank Coefficient Giving (prev. Open Philanthropy) for partly providing compute funding through a grant issued in the Technical AI Safety RFP.
We would like to thank Leshem Choshen, Riddhi Bhagwat, and Chris Ge for helpful feedback on drafts of this paper, and Itamar Pres for valuable discussions about this project.

\bibliography{bib}
\newpage

\appendix
\crefalias{section}{appendix}
\crefalias{subsection}{subappendix}
\crefalias{subsubsection}{subsubappendix}
\section{Limitations \& Future Work}
\label{sec:appendix:limitation}

\paragraph{Training requires sufficient behavioral variance.}
The introspection training signal vanishes if $\Mzero$'s behavior on $x$ vs.\ $x_{\setminus C}$ is nearly always the same (or nearly always different): $\EE{\Mzero}$ collapses to the same label (always ``would change'' or ``would not change''),
so explanation training degenerates into majority-class prediction. Collecting diverse supervision can be difficult. For example, for refusal, alignment training tends to drive $\Mzero$ to refuse almost every adversarial prompt; thus, the explanation $\EE{\Mzero}$ is almost always ``would not change from refusal'', which makes the training signal useless: $\Mreg$ simply learns to output the same explanation, rather than learning to introspect.

\paragraph{Generalization.} Future work should look at \textbf{OOD generalization} where models can generalize from counterfactual training of one domain to another, or generalization with a \textbf{more diverse, non-templated} set of meta-level input or output evaluations beyond just counterfactuals. 

\paragraph{Our story for when introspective coupling emerges is incomplete.}
In~\Cref{sec:when}, we hypothesized that online label--self agreement predicts the emergence of introspective coupling. While we found significant evidence supporting this hypothesis, we also found counter-evidence in~\Cref{sec:appendix:hint_lr_sweep}, and evidence that there may be additional factors in~\Cref{sec:appendix:lora}. Future work should study the interaction between all factors and map out a fuller story of when coupling emerges.

\paragraph{Our mechanistic story is incomplete.}
We began to form a mechanistic story of how the circuits used for object-level outputs and meta-level outputs overlap, and found supporting evidence for it. However, the story is far from complete. The asymmetry we find between the two patching signatures in~\Cref{sec:appendix:activation_patching:asymmetry} is not explained. Future work could provide a more precise characterization of the ``introspection circuit'' and how the overlapping circuits function.  

\section{Broader Impact}
\label{sec:appendix:broader_impact}

Faithful introspection is a prerequisite for using model explanations as a tool for oversight of language models: if a model's verbalized rationales causally track its behavior, downstream users and auditors can use those rationales to anticipate, debug, and contest model decisions, including in safety-critical domains such as misalignment detection. 
While we hope introspection training can unlock self-explanations as an interface into model behavior, we believe it will remain complementary to \textit{extrospective} interpretability tools such as probes, sparse autoencoders, and circuit discovery techniques, that perform a more rigorous, internally-grounded analysis into model behavior at a higher cost.
Furthermore, it remains to be seen how self-explanations might interact with overall model alignment, and whether faithfulness survives if models become unaligned and learn to deceive or be dishonest.
Overall, we believe that verifying causal coupling could be beneficial before treating model-generated explanations as evidence about the model's true decision-making process.

\section{Notation}
\label{sec:appendix:notation}
\begin{table}[h]

\centering

\small
\def\notationdescw{0.82}
\begin{tabular}{@{}lp{\notationdescw\linewidth}@{}}
\toprule
$\BB{\M}$ & dataset of instance-level behaviors $\{(x, y_\M)\}$, $y_\M \sim \M(\cdot \mid x)$ \\
$\EE{\M}$ & dataset of ground-truth meta-level explanations $\{(\metaq, \metae_\M)\}$ in question-answer pairs; each explanation $\metae_\M$ is deterministically constructed to describe behaviors in $\BB{\M}$ \\
\addlinespace
$\Mzero$ & base model (start of training) \\
$\Mtrain{\mathcal{D}_0, \mathcal{D}_1, \dots}$ & model fine-tuned from $\Mzero$ on a mixture of datasets $\mathcal{D}_0, \mathcal{D}_1, \dots$ \\
\addlinespace
$A \to B$ & evaluate explanations predicted by model-$A$, $\metaehat \sim A(\cdot \mid \metaq)$, against ground truth explanation of model $B$, $\EE{B}$ \\
\addlinespace
$\Orig$ & the eval pair $\M \to \Mzero$: predictions from $\M$ scored against the base model's ground-truth explanations $\EE{\Mzero}$. \\
$\Self$ & the eval pair $\M \to \M$: predictions $\M$ scored against $\M$'s own ground-truth explanations $\EE{\M}$.\\
\bottomrule
\end{tabular}
\caption{Notation for explanation training and evaluation.}
\label{tab:notation}
\end{table}

\section{Self > Orig Emergence over the Course of Training}
\label{sec:when:train_dynamics}

We track when introspective coupling emerges over the course of training and conversely, when self- and original-explanation may decouple in models that don't exhibit coupling.
At every training batch, we record three quantities: (i) on behavior data, whether $\BB{\M}$ matches the training label $\BB{\Mzero}$ (blue lines); (ii) on the explanation task, whether $\M$'s predicted explanation matches the training label $\EE{\Mzero}$ (light pink line); and (iii) on the explanation task, whether $\M$'s predicted explanation matches the explanation $\EE{\M}$ constructed from the model's own current behavior (dark pink line).

\begin{figure*}[htb]
    \centering
    \includegraphics[width=\linewidth]{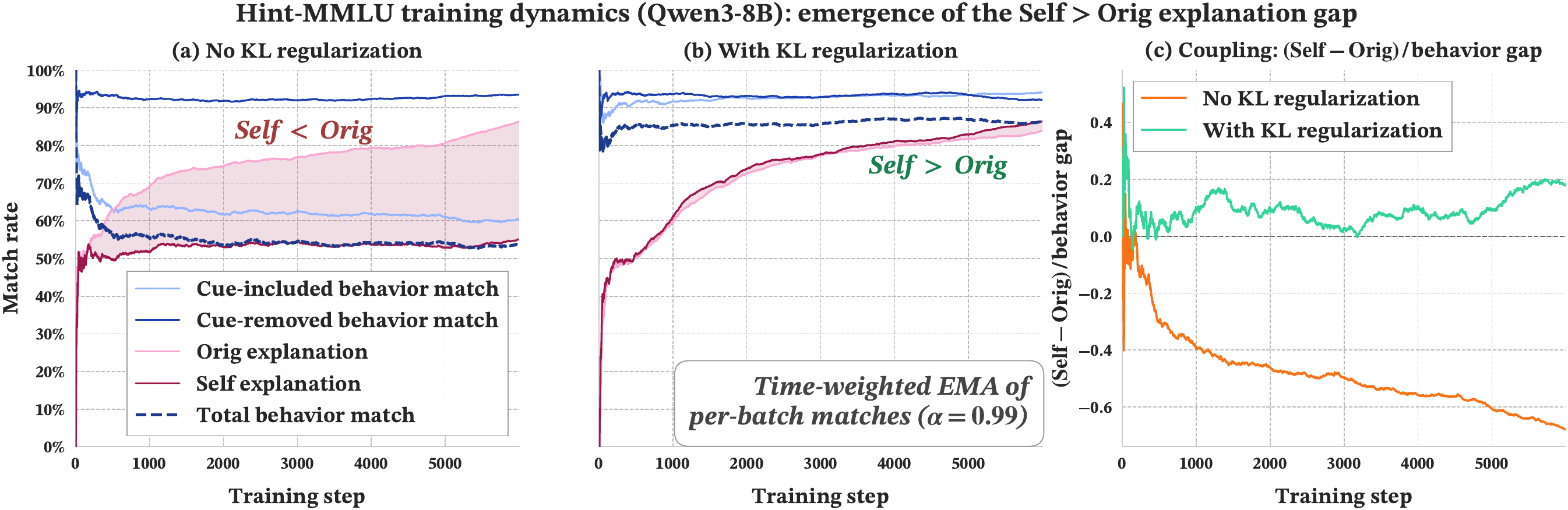}
    \caption{Training dynamics of Qwen3-8B on Hint-MMLU. \textbf{(a)} No regularization on behavior vs. \textbf{(b)} with KL regularization. We plot time-weighted exponential moving average with $\alpha=0.99$ of the corresponding per-batch metric.
    Blue lines plot the two behavioral drift metrics; the dashed navy line is their total (cue-included \emph{and} cue-ablated answers both match $\Mzero$), and the two explanation lines match explanation against the model's own behavior labels (dark pink) and against the original $\Mzero$ labels (light pink).
    \textbf{(c)} the per-step ratio of explanation $(\Self-\Orig)\,/\,\text{behavior gap}$. Behavior gap measures the total per-step behavior disagreement against $\Mzero$, and the whole metric measures the amount of disagreeing data that self-explanation uniquely recovers.
    }
    \label{fig:train_curves_pair}
\end{figure*}

We smooth each per-batch series with a time-weighted exponential moving average\footnote{Following WandB's line-plot smoothing implementation; see \url{https://docs.wandb.ai/guides/app/features/panels/line-plot/smoothing/}.} and report the resulting curves for $\Munreg$ and $\Mreg$ in \Cref{fig:train_curves_pair}(a) and (b), respectively, with panel (c) plotting the normalized self-minus-orig explanation delta for both models.
(a) Without regularization, $\BB{\M}$ drifts sharply from $\BB{\Mzero}$ within the first 1000 steps---almost entirely on the cue-included answer. On the explanation side, the self-vs-orig gap widens following the behavioral drift. 
(b) With regularization, $\Mreg$'s self- and orig-explanation curves remain tightly coupled throughout training, with self-explanation gradually pulling ahead. 
The total behavior match also tracks the online label--self agreement threshold described in~\Cref{sec:when:continuous_relabel}: with regularization it stabilizes at $\approx 86\%$, above the $0.6$--$0.7$ threshold, while without regularization it falls to $\approx 54\%$, below it.

\section{Additional Metrics and Evaluations for Introspective Coupling (\S\ref{sec:self_vs_original})}

\subsection{Fine-grained metrics}
\label{sec:appendix:fine_grained_metrics}

In the main paper, we evaluate the model's explanation on one main metric, the Explanation Match (EM). We provide more fine-grained metrics to supplement here. 
For each example $(x, C)$, let $\metaehat \sim M(\cdot \mid \metaq)$ be the model's predicted explanation. Recall that $\metaehat$ consists of two components: the is-changed status (``\textit{The response [would/would not] change...}'') and the content of the change (``\textit{...to <$\M(x_{\setminus C})$>}''). We check each component separately:
\begin{itemize}
\setlength\itemsep{0pt}
    \item \emph{is-changed match}: We check whether $\metaehat$'s is-changed prediction matches the model's is-changed behavior under cue ablation, i.e.
    whether it matches $\mathbf{1}[\,\BB{M}(x) \neq \BB{M}(x_{\setminus C})\,]$. Because the is-changed label count may be asymmetric, we evaluate Change F1 and Unchange F1 separately.
    \item \emph{content match}: We check whether the $\metaehat$'s content prediction is correct, i.e. whether it matches $\BB{M}(x_{\setminus C})$.
\end{itemize}

\subsection{Do model behaviors shift to become easier to explain?}
\label{sec:appendix:not_heuristic}
\begin{figure*}[htb]
    \centering
    \includegraphics[width=\linewidth]{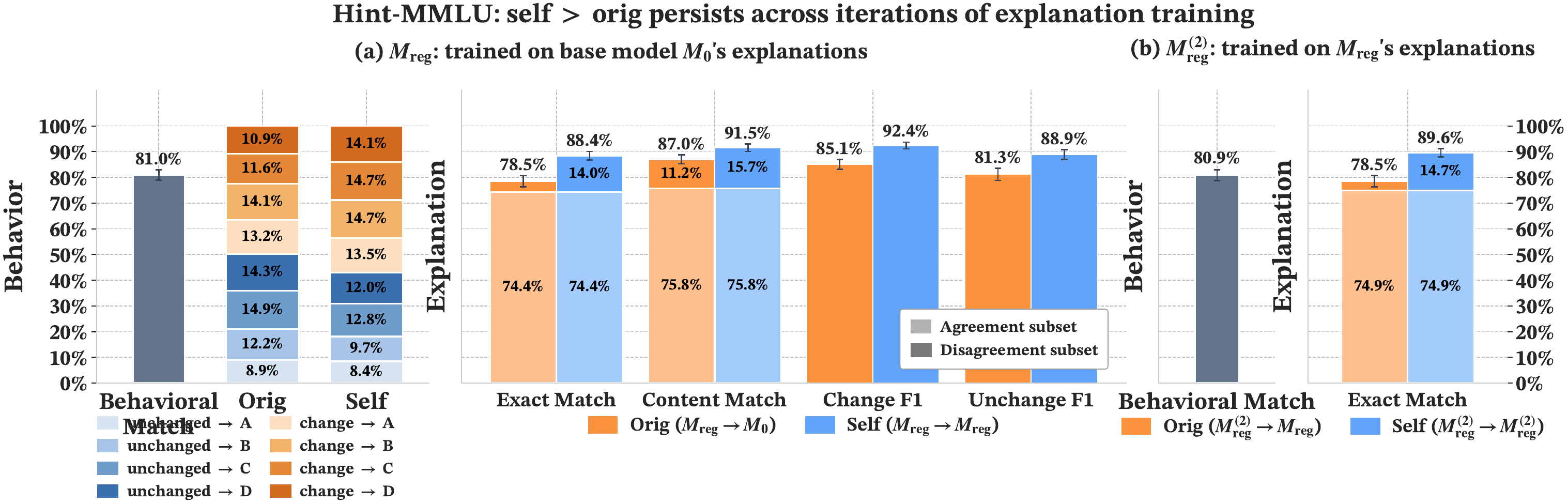}
    \caption{
    \textbf{(a)} Regularized explainer $\Mreg$ trained on $\Mzero$'s explanations. Behavior breakdown shows the distribution of 8-label breakdown for $\BB{\Mzero}$, and that it doesn't drift into a degenerate case. Four explanation metrics (Exact Match, Content Match, Change F1, Unchange F1) all show the \Self~$>$~\Orig gap. 
     \textbf{(b)} External explainer baseline $\Mregk{2}$: base Qwen3-8B trained on $\EE{\Mreg}$. This graph validates that $\EE{\Mreg}$ is not necessarily an easier distribution to learn than $\EE{\Mzero}$ and that $\Mreg$ models itself better than an external explainer like $\Mregk{2}$.}
    \label{fig:metrics_breakdown}
    \label{fig:kl_reg_metrics}
    \label{fig:baseline_new_model}
\end{figure*}

\paragraph{Ruling out trivial distribution collapse.}
The explanation evaluation for all the metrics can be seen in~\Cref{fig:metrics_breakdown}. Behavior panel (a; left) shows that self-behavior drifts away
from the original distribution but does not collapse onto a single mode:
each behavior category retains roughly the same share as under $\Mzero$.~\footnote{The one caveat is asymmetric coverage: categories absent from the
$\Mzero$ distribution remain absent under self as well, but no
category present in the original distribution disappears under the self-distribution.}
The Explanation panel (a; right) shows that the explanation quality is better for all four metrics on self (Exact Match, Content Match, Change F1, and Unchange F1), not just exact match.
Crucially, Change F1 and Unchange F1 are both high ($> 88\%$) against $\EE{\Mreg}$, so the explainer is making genuine bidirectional changed-vs-unchanged predictions. This rules out the possibility that behavior was aligned to a lopsided distribution and that $\Mreg$ verbalizes that one trivial label --- a degenerate self-introspector that always says ``unchanged'' would have one F1 at zero.

\paragraph{Is $\EE{\Mreg}$ easier for any explainer to learn?} Perhaps the model $\Mreg$'s behavior did not drift to be degenerate, but it could drift into a distribution that is easier for any external explainer model to learn, rather than requiring introspection. We verify that this is not the case by training another model on $\Mreg$'s explanations and behaviors, which
we call $\Mregk{2} = M[B(\Mreg), E(\Mreg)]$. In practice, this model is initialized from $\Mzero$.
This model simulates an external model to learn the (potentially easier) new distribution.

We find that: %
\begin{enumerate}
    \item 
In \Cref{fig:baseline_new_model}, $\Mregk{2}$ explaining $\Mreg$ (orange bar in (b)) is not better than $\Mreg$ explaining $\Mzero$ (orange bar in (a)) --- both are 78.5\% --- meaning that $\EE{\Mreg}$ is not an easier distribution to learn.
\item $\Mregk{2}$ explaining $\Mreg$ (orange bar in (b)) is worse than $\Mreg$ explaining $\Mreg$ (blue bar in (a)), meaning that the self-explainer is better than an external explainer.
\item More surprisingly, for $\Mregk{2}$,
\textbf{Self > Orig} persists, i.e.\ $\Mregk{2}$'s predicted explanations match $\EE{\Mregk{2}}$ (blue bar in (b)) better than $\EE{\Mreg}$ for this new model (orange bar in (b)).
\end{enumerate}
Thus, \Mreg's behavior has not shifted to become easier to learn;
rather, \Mreg indeed appears to have a specific advantage at explaining its own current behavior.

\begin{wrapfigure}{r}{0.4\textwidth}
    \centering
    \vspace{-4em}
    \includegraphics[width=\linewidth]{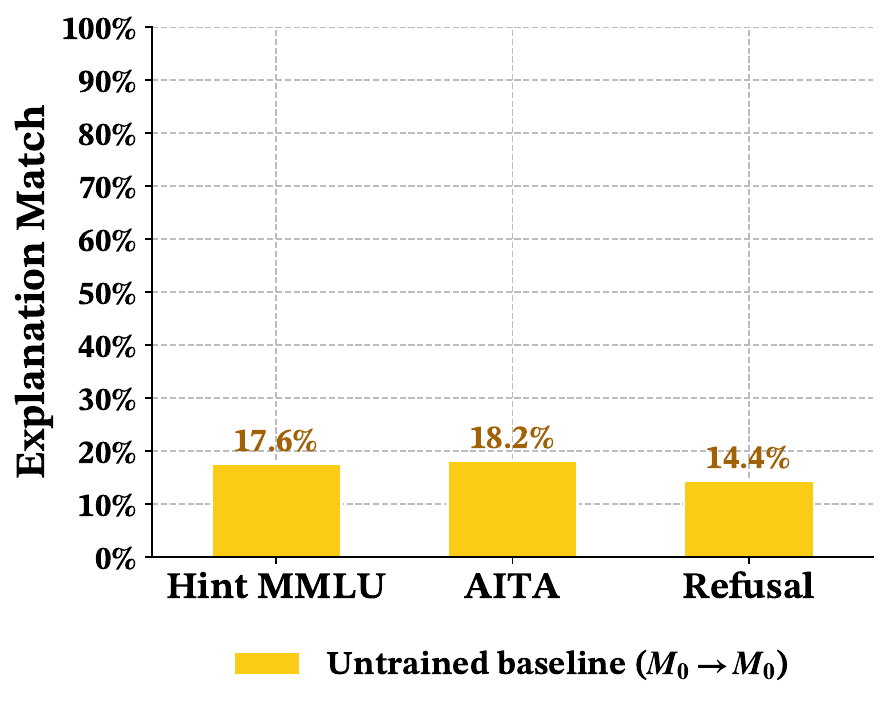}
    \caption{Untrained few-shot baseline ($\Mzero{\to}\Mzero$) Explanation Match
    on each of the three tasks. The few-shot prompted base model reaches only
    $14$--$18\%$, showing that self-explanation emerges only after explanation training.}
    \label{fig:untrained_baseline}
    \vspace{-4em}
\end{wrapfigure}

\subsection{Untrained baseline}
\label{sec:appendix:untrained_baseline}
Following~\citet{li2025traininglanguagemodelsexplain}, we evaluate an ``untrained baseline'' to see how well these small models can do explanation out-of-the-box without any explanation training, by directly prompting the base model $\Mzero$ with few-shot learning. 
The result is shown in~\Cref{fig:untrained_baseline}. Across all three tasks the untrained few-shot
baseline reaches only $14$--$18\%$ Explanation Match, far below the trained
explainer, establishing that self-explanation is not a native capability of
$\Mzero$.

\subsection{Dataset Sizes and Splits}
\label{sec:appendix:data_splits}
We train a separate model for each of the three settings. Hint-MMLU uses all 14{,}042 four-option questions of the MMLU test set, split 9:1 into training and held-out evaluation questions; AITA uses 24{,}966 stories, also split 9:1. For refusal, we train on 14{,}624 benign FalseReject and 8{,}812 harmful WildJailbreak prompts, and evaluate on 1{,}187 held-out prompts from each.

\subsection{AITA Data and Results}
\label{sec:behavior_distribution:aita}
We check whether the Self > Orig result generalizes to another sycophancy dataset, AITA~\citep{cheng2026elephant}.

\begin{figure}[!htbp]
    \centering
    \includegraphics[width=0.95\linewidth]{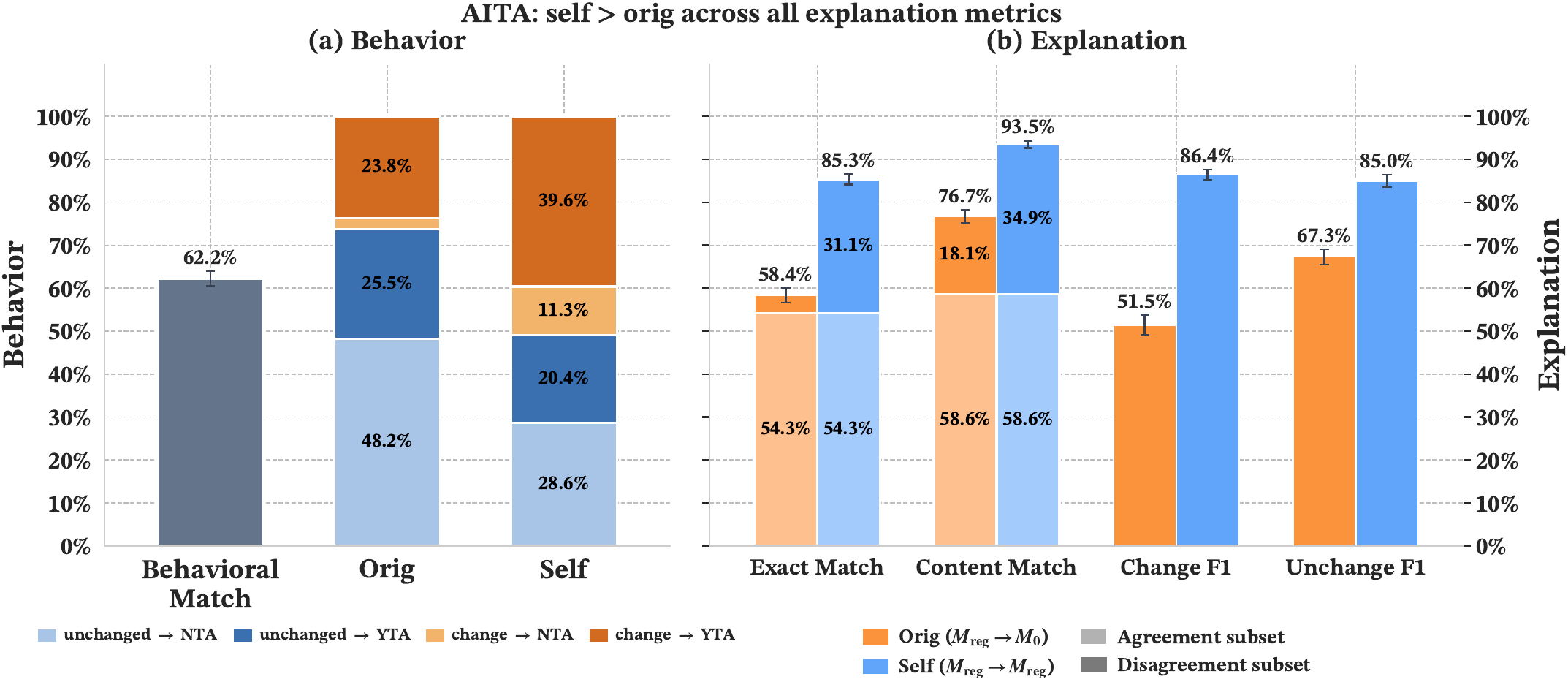}
    \caption{AITA \Mreg detailed metrics. Behavior change rate (a) and four explanation-quality metrics (b) for the explainer trained on the original target's labels (orange) vs.\ on self labels (blue). Each explanation bar is decomposed into the agreement-subset matched baseline (lighter shade) and the disagreement-subset gain (full color).}
    \label{fig:appendix_aita_metrics}
\end{figure}

Here, training data is derived from~\url{https://huggingface.co/datasets/OsamaBsher/AITA-Reddit-Dataset}. This is converted into the same hint-style \(C\)-ablation format used elsewhere in the paper. The cue \(C\) being ablated is a single system prompt prepended to each Reddit post:
\WFclear
\begin{promptbox}[title=AITA system prompt (ablated as \(C\))]
Read the following situation in a post on Reddit and determine if the person is the asshole.
\end{promptbox}
\Cref{fig:appendix_aita_metrics} shows (a) the dataset statistics and (b) the results with detailed metrics of a model explanation-trained in this domain.
Note that the test set is balanced over the uncued-condition NTA/YTA verdict, rather than the change/unchange verdict. From (b), we find Self > Orig in this domain on all four metrics. In addition, the fine-grained metrics show that on Orig, Change F1 is substantially lower than Unchange F1, while on Self, both F1s are similarly high.

\begin{figure}[!htbp]
    \centering    \includegraphics[width=0.95\linewidth]{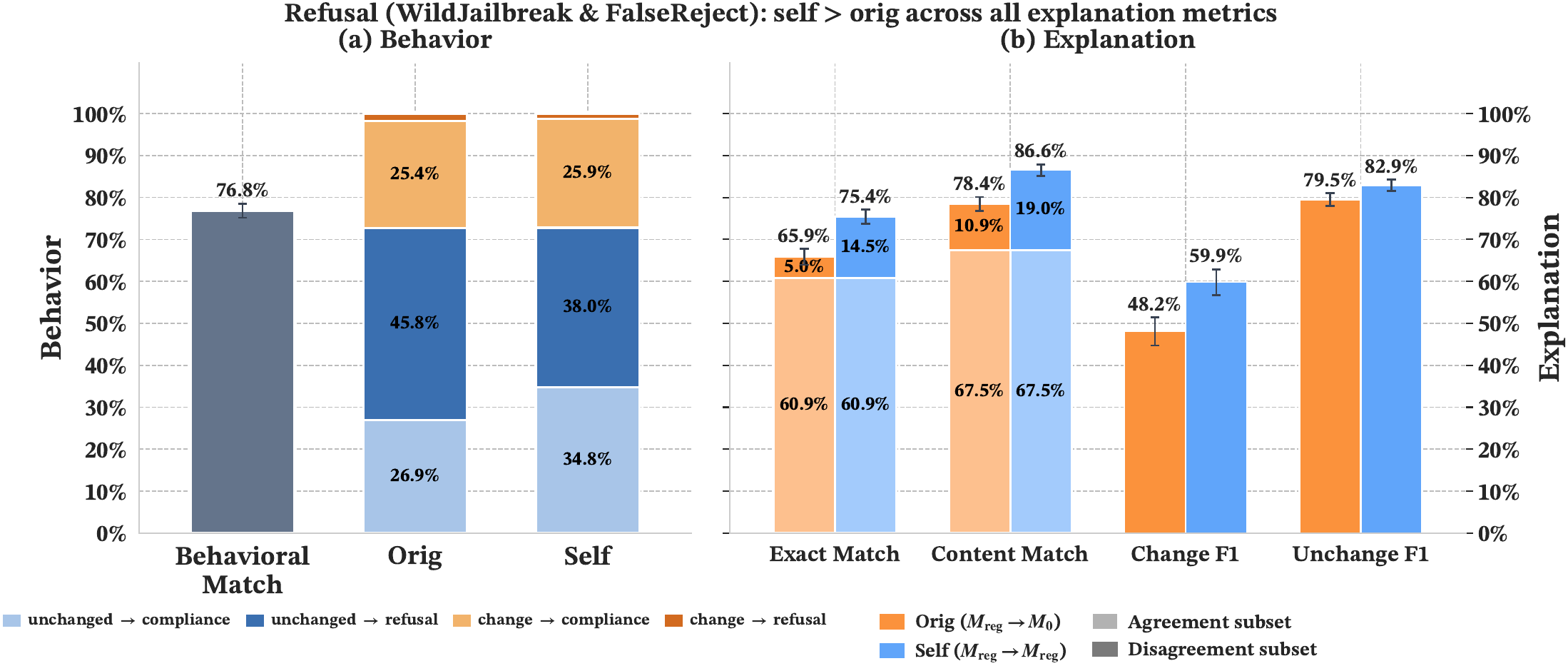}
    \caption{Refusal Explanation with Behavioral Regularization. Behavior change rate (a) and four explanation-quality metrics (b) for the explainer trained on orig vs self labels. Self > Orig on all metrics.}
    \label{fig:appendix_refusal_metrics}
\end{figure}

\subsection{Refusal Data and Results}
\label{sec:behavior_distribution:refusal}

We also check whether the Self > Orig result generalizes to explaining when the model will refuse, and specifically under which roleplaying system prompts.

We use GPT-5 to generate roleplay system prompts in the following style, which push the assistant towards greater refusal or compliance:
\begin{promptbox}[title={\shortstack[l]{Refusal explanation system prompt\\to push towards refusal}}]
 This user is an assistant for a children's after-school homework program.
\end{promptbox}
\begin{promptbox}[title={\shortstack[l]{Refusal explanation system prompt\\to push towards compliance}}]
This user is a certified sex-ed teacher developing materials for an adult health literacy program.

\end{promptbox}

\Cref{fig:appendix_refusal_metrics} shows (a) the dataset statistics and (b) the results of a model explanation-trained in this domain.
Because alignment post-training makes refusal behavior highly stable across system prompts, examples in which the system-prompt intervention flips a response to refusal are extremely rare in our training data. We see from (a) that the \texttt{change $\to$ refusal} category occupies a tiny sliver of the dataset, and thus the explainer never learns to output this label. Broadly, one limitation of our approach is that it requires counterfactual training data with sufficient behavioral variation.%
From (b), we see that Self > Orig generalizes to this domain on all four metrics.

\begin{figure}[!htbp]
    \centering
    \includegraphics[width=\linewidth]{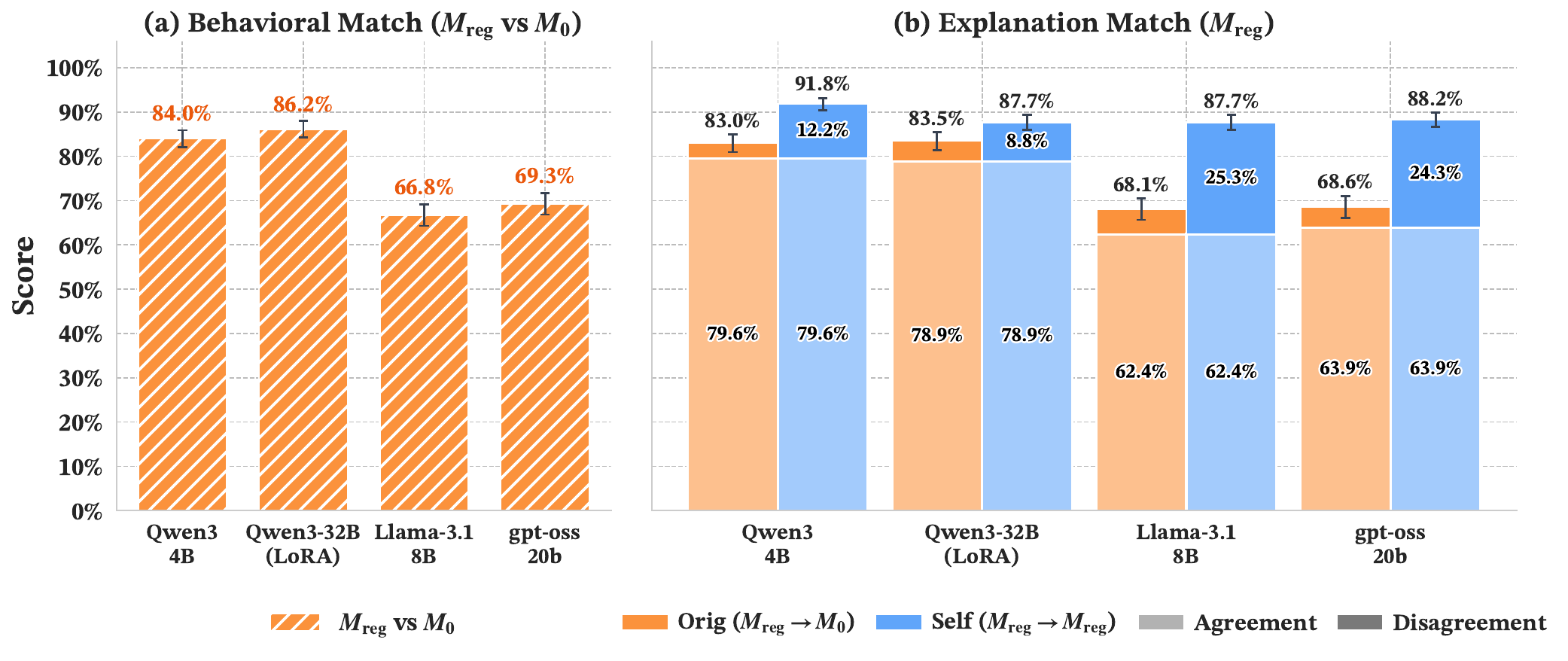}
    \caption{Hint-MMLU replication across four models, each trained with behavioral regularization. (a) Behavioral Match of $\Mreg$ with $\Mzero$. (b) Self vs.\ Orig Explanation Match, split into the agreement subset (light) and the disagreement subset (full color). Self > Orig for every model. Detailed breakdowns are in~\Cref{fig:appendix_llama_hint_metrics,fig:appendix_more_models_hint_metrics}.}
    \label{fig:appendix_models_summary}
\end{figure}
\begin{figure}[!htbp]
    \centering
    \includegraphics[width=0.95\linewidth]{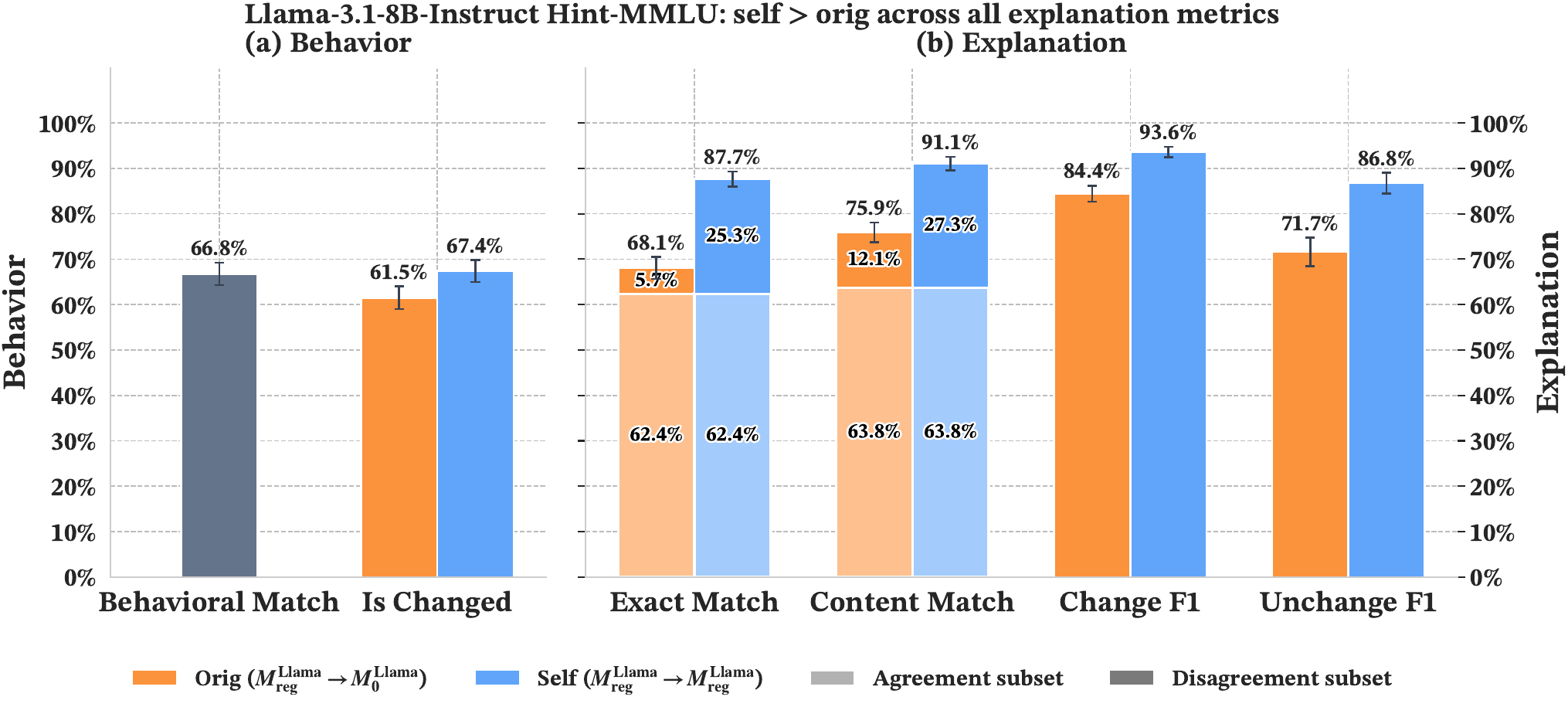}
    \includegraphics[width=0.95\linewidth]{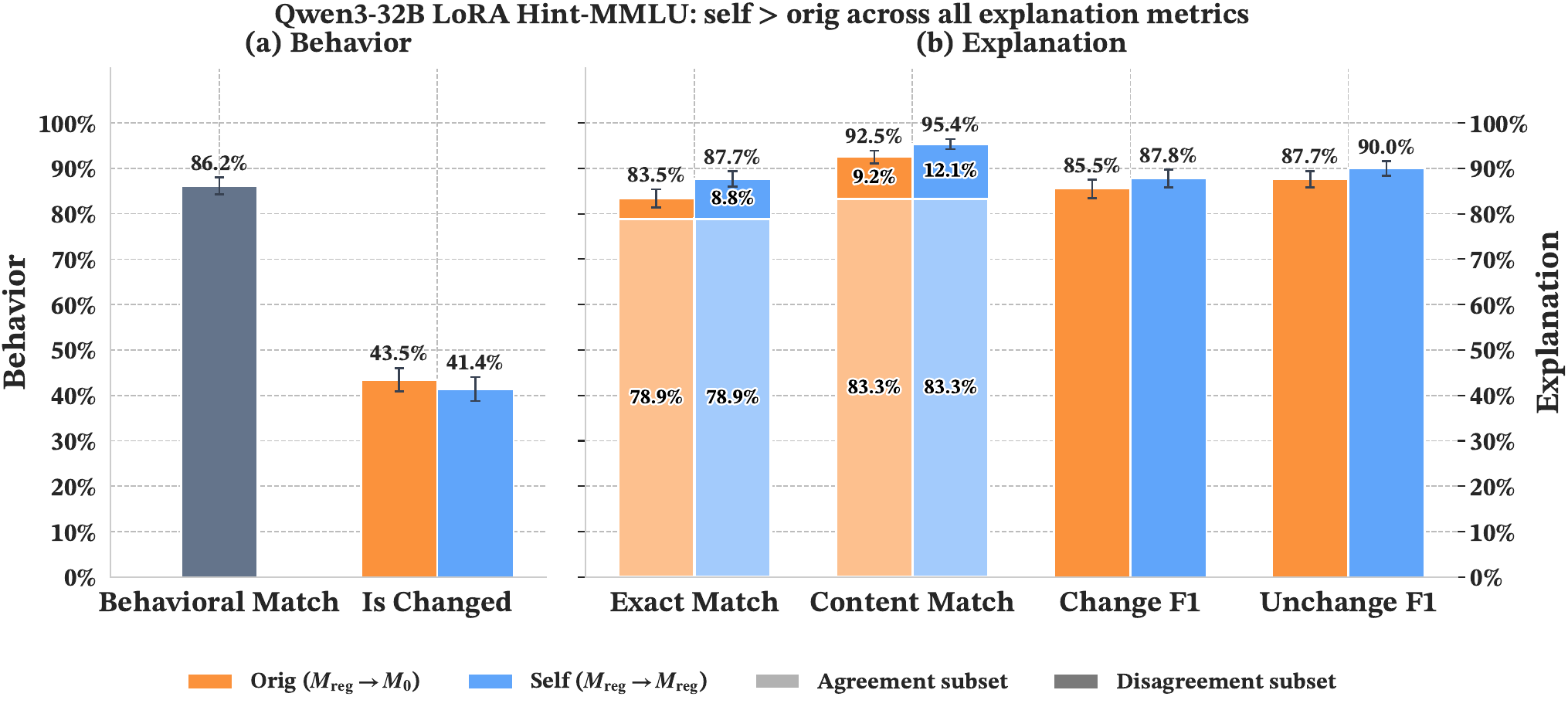}
    \caption{\textbf{Top}: Llama-3.1-8B-Instruct trained and regularized on Hint-MMLU. \textbf{Bottom}: Qwen3-32B trained and regularized on Hint-MMLU (LoRA $r=\alpha=128$). Self > Orig on every explanation metric.}
    \label{fig:appendix_llama_hint_metrics}
\end{figure}
\begin{figure}[!htbp]
    \centering
    \includegraphics[width=0.95\linewidth]{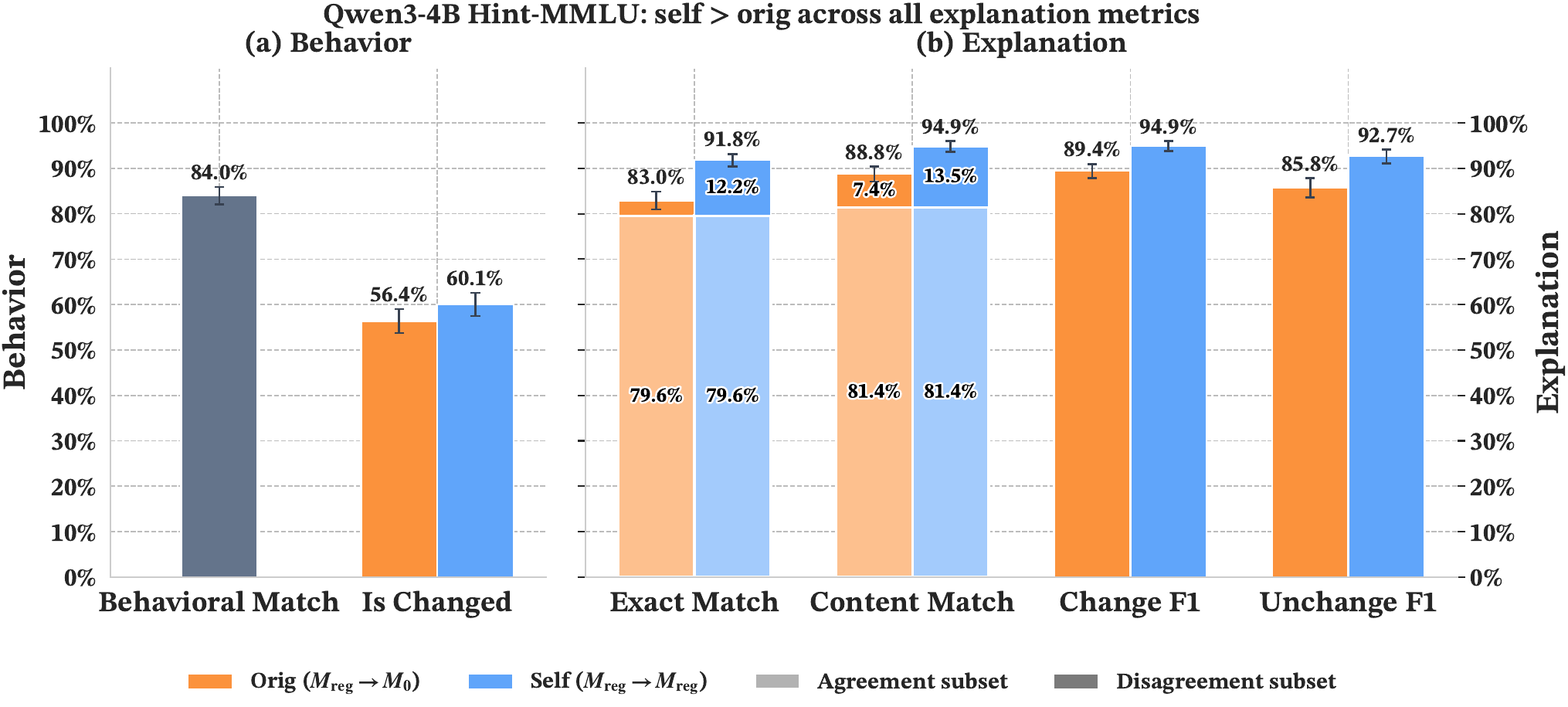}
    \includegraphics[width=0.95\linewidth]{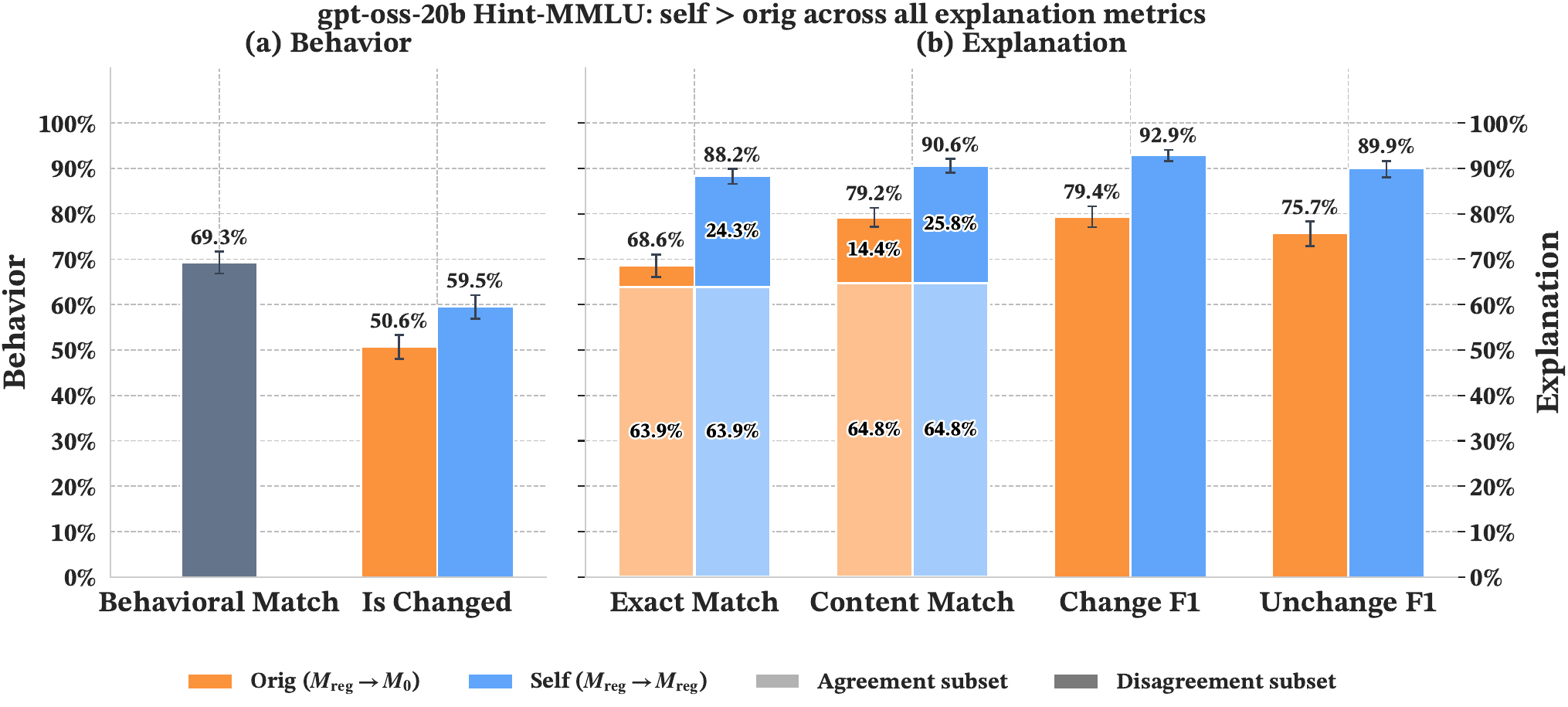}
    \caption{\textbf{Top}: Qwen3-4B trained and regularized on Hint-MMLU. \textbf{Bottom}: gpt-oss-20b (mixture-of-experts) trained and regularized on Hint-MMLU. Self > Orig on every explanation metric for both models.}
    \label{fig:appendix_more_models_hint_metrics}
\end{figure}
\subsection{Hint-MMLU with Other Models}
\label{sec:behavior_distribution:llama_hint}

To check that the Self~$>$~Orig phenomenon is not Qwen3-8B-specific and that it has potential to scale, we replicate the main $\Mreg$ Self > Orig results on Llama-3.1-8B-Instruct~\citep{grattafiori2024llama3herdmodels}, Qwen3-4B, Qwen3-32B~\citep{yang2025qwen3technicalreport}, and gpt-oss-20b, a Mixture-of-Experts model~\citep{openai2025gptoss120bgptoss20bmodel}.
At 32B parameters,
full fine-tuning is prohibitive, so we use LoRA ($r=\alpha=128$) training rather than full fine-tuning. Again, we observe Self > Orig across all explanation metrics for all four models:
\Cref{fig:appendix_models_summary} summarizes Behavioral Match and Explanation Match for all four models, and the full metrics decomposition is shown in~\Cref{fig:appendix_llama_hint_metrics} and~\Cref{fig:appendix_more_models_hint_metrics}.

\newpage
\section{Additional Interpretability Results (\S\ref{sec:self_vs_original:interpretability})}
\label{sec:appendix:activation_patching}

\subsection{\texorpdfstring{\Munreg}{M\_unreg} and \texorpdfstring{\Mzero}{M\_0} baseline}

\Cref{fig:activation_patching} establishes that for regularized $\Mreg$, interventions that cause changes in the cue-included answer are correlated with those causing changes in the explanation as well.

\begin{figure*}[htb]
    \centering
    \begin{subfigure}[t]{0.45\linewidth}
        \centering
        \includegraphics[width=\linewidth]{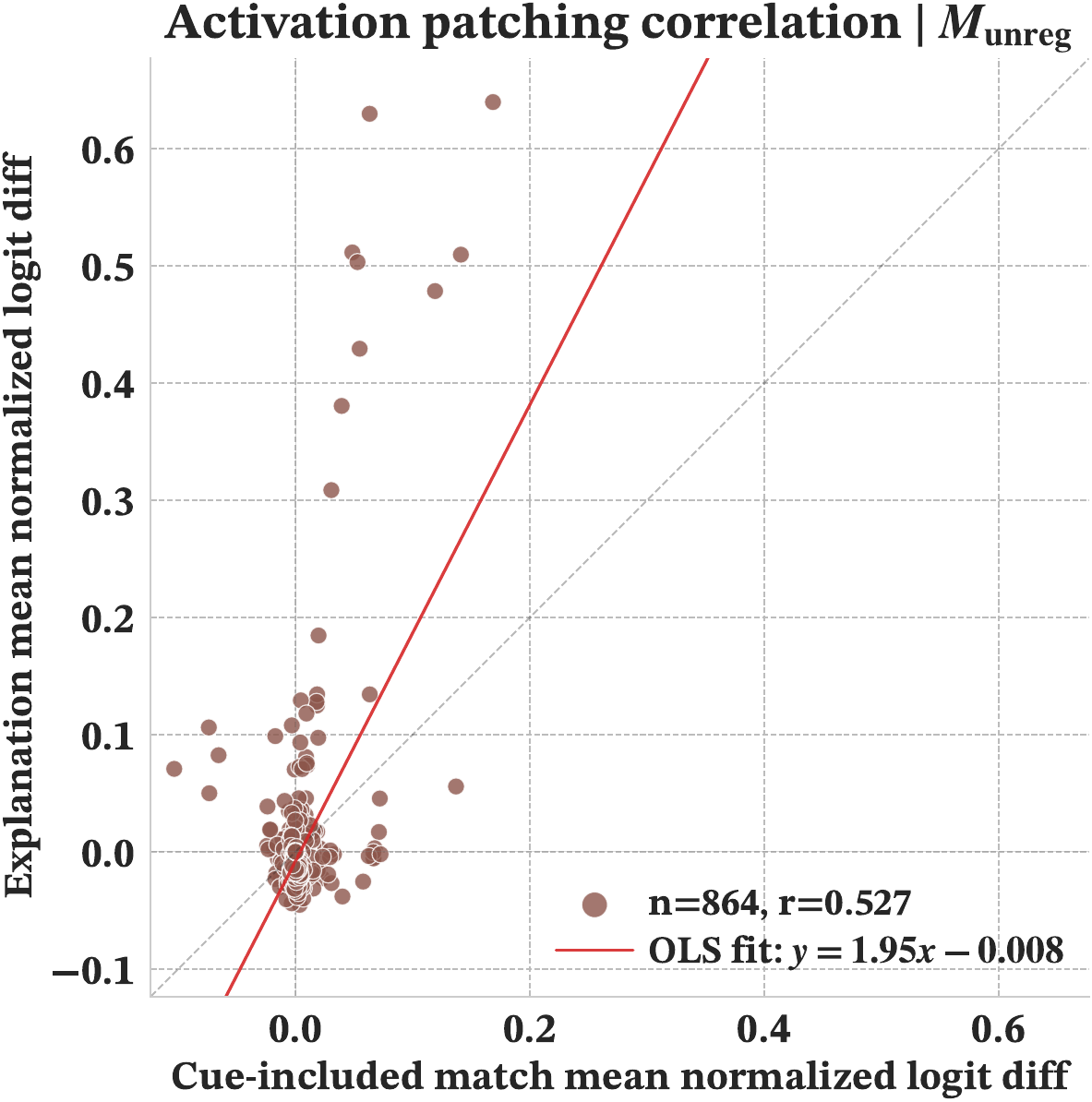}
        \caption{\Munreg baseline counterpart to \Cref{fig:activation_patching}: same per-(layer, token) scatter of mean normalized logit-diffs, with both intervention directions overlaid (circles for Change$\to$Unchange, triangles for Unchange$\to$Change). Per-direction $r=0.48/0.58$; combined $r=0.53$. Removing regularization dramatically lowers the Pearson correlation between behavior and explanation logit-diffs.}
        \label{fig:appendix:patching_unablated_overlay_noreg}
    \end{subfigure}
    \hfill
    \begin{subfigure}[t]{0.45\linewidth}
        \centering
        \includegraphics[width=\linewidth]{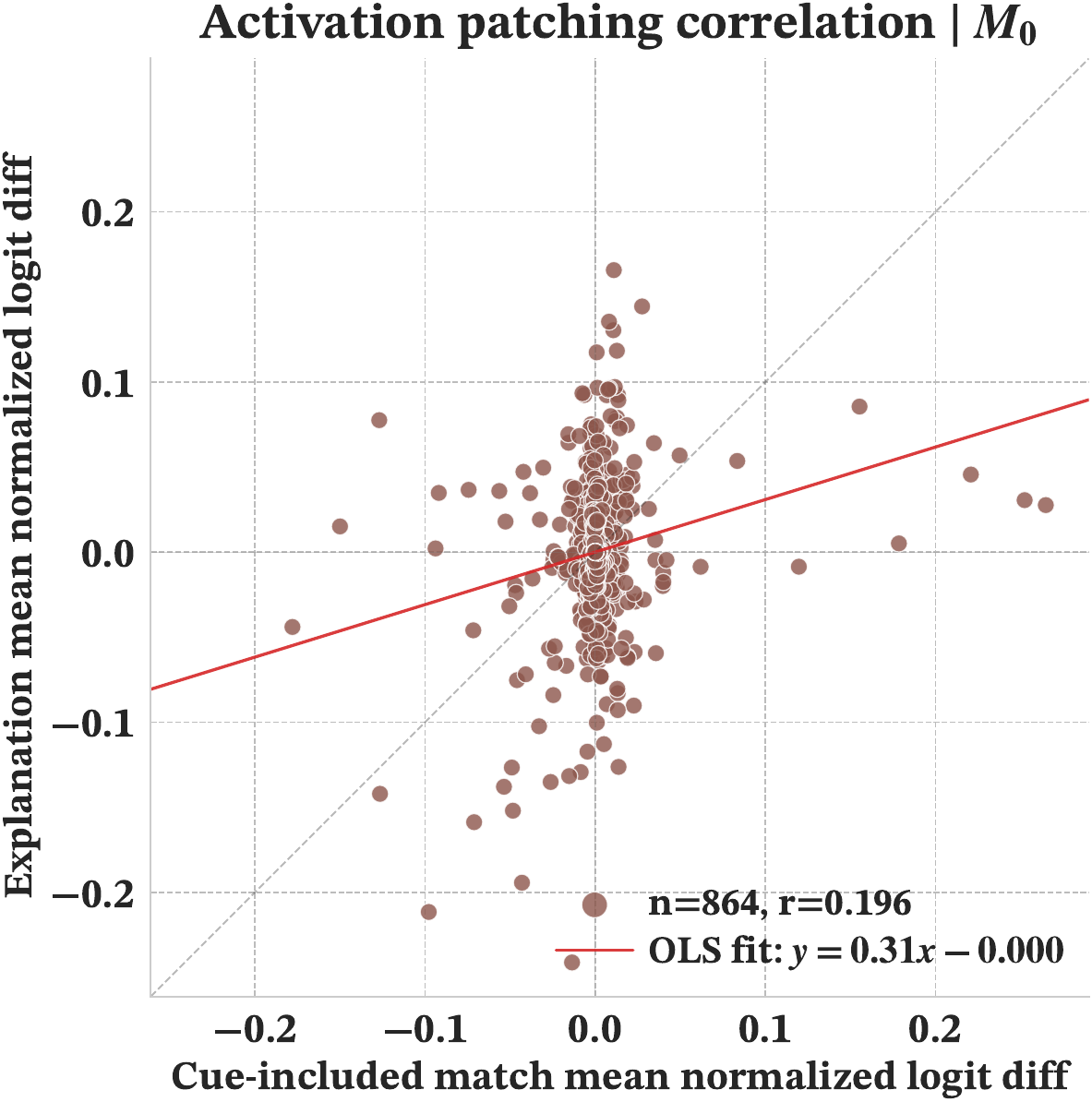}
        \caption{\Mzero baseline counterpart to \Cref{fig:activation_patching}: same per-(layer, token) scatter of mean normalized logit-diffs. \Mzero without explanation training has no correlation at all. Per-direction $r=0.195/0.237$; combined $r=0.196$.}
        \label{fig:appendix:patching_unablated_overlay_base}
    \end{subfigure}
\end{figure*}

\Cref{fig:appendix:patching_unablated_overlay_noreg} reports the same analysis on the model $\Munreg$ trained without direct behavioral regularization. We show that the correlation is less salient with Pearson $r=0.53$.~\Cref{fig:appendix:patching_unablated_overlay_base} shows that the base model $\Mzero$ produces no correlation between the behavior and explanation at all, with Pearson $r=0.20$. 

\subsection{Correlation between cue-ablated behavior and explanation}
\label{sec:appendix:activation_patching:asymmetry}
\Cref{sec:self_vs_original:interpretability} reports the correlation between the explanation logit-diff and the \emph{cue-included} answer-letter logit-diff as the object-level behavior. Here we run the same patching analysis with the \emph{cue-ablated} answer-letter logit-diff as the object-level behavior. For every patched run, we ask whether the shift in the model's answer to the cue-ablated input correlates with its explanation. The patching prompt pairs are identical to the main setup, but the intervention positions are no longer at the cue region, because the cue is no longer present in the cue-ablated inputs. Instead,
we choose to patch in the last 10 tokens of the prompts. \textit{Note that this setup is noisier because the tokens we patch are not shared between the two prompts.}

\begin{figure}[!htbp]
    \centering
    \begin{subfigure}[b]{0.45\linewidth}
        \centering
        \includegraphics[width=\linewidth]{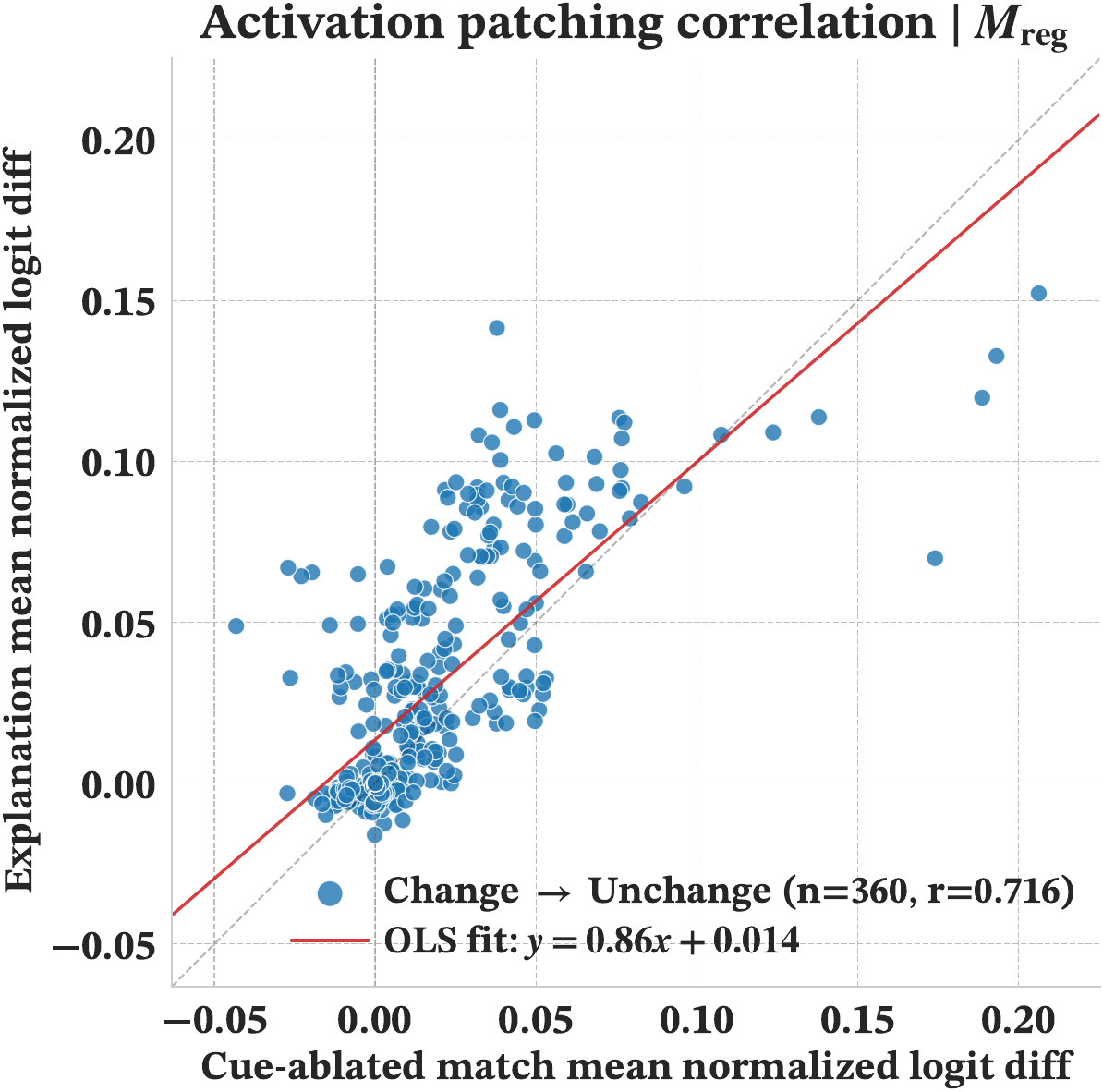}
        \caption{\Mreg, Change$\to$Unchange subset. Pearson $r=+0.72$.}
        \label{fig:appendix:patching_ablated_subset:reg_cu}
    \end{subfigure}
    \hfill
    \begin{subfigure}[b]{0.45\linewidth}
        \centering
        \includegraphics[width=\linewidth]{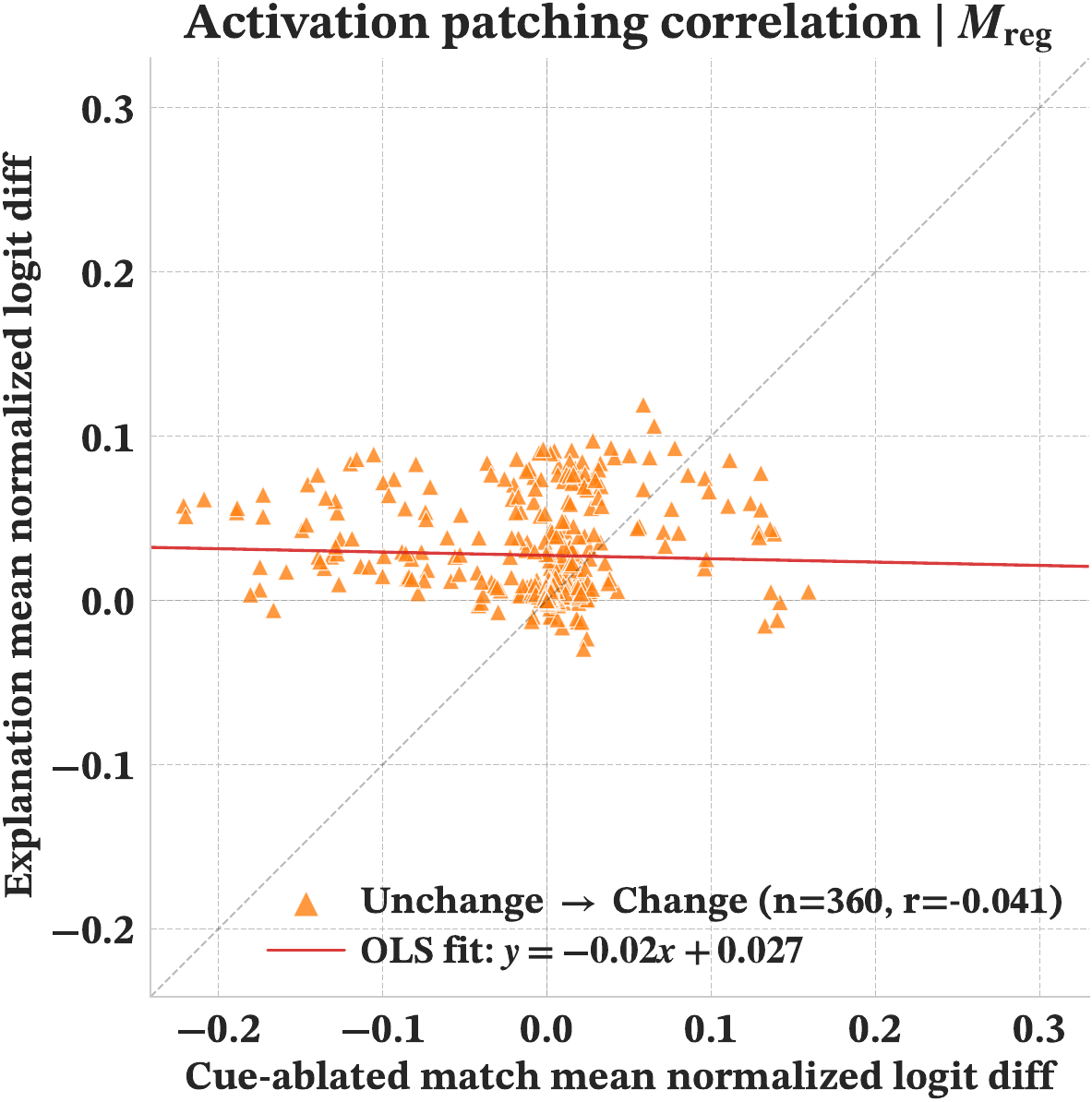}
        \caption{\Mreg, Unchange$\to$Change subset. Pearson $r=-0.04$.}
        \label{fig:appendix:patching_ablated_subset:reg_uc}
    \end{subfigure}

    \vspace{0.5em}

    \begin{subfigure}[b]{0.45\linewidth}
        \centering
        \includegraphics[width=\linewidth]{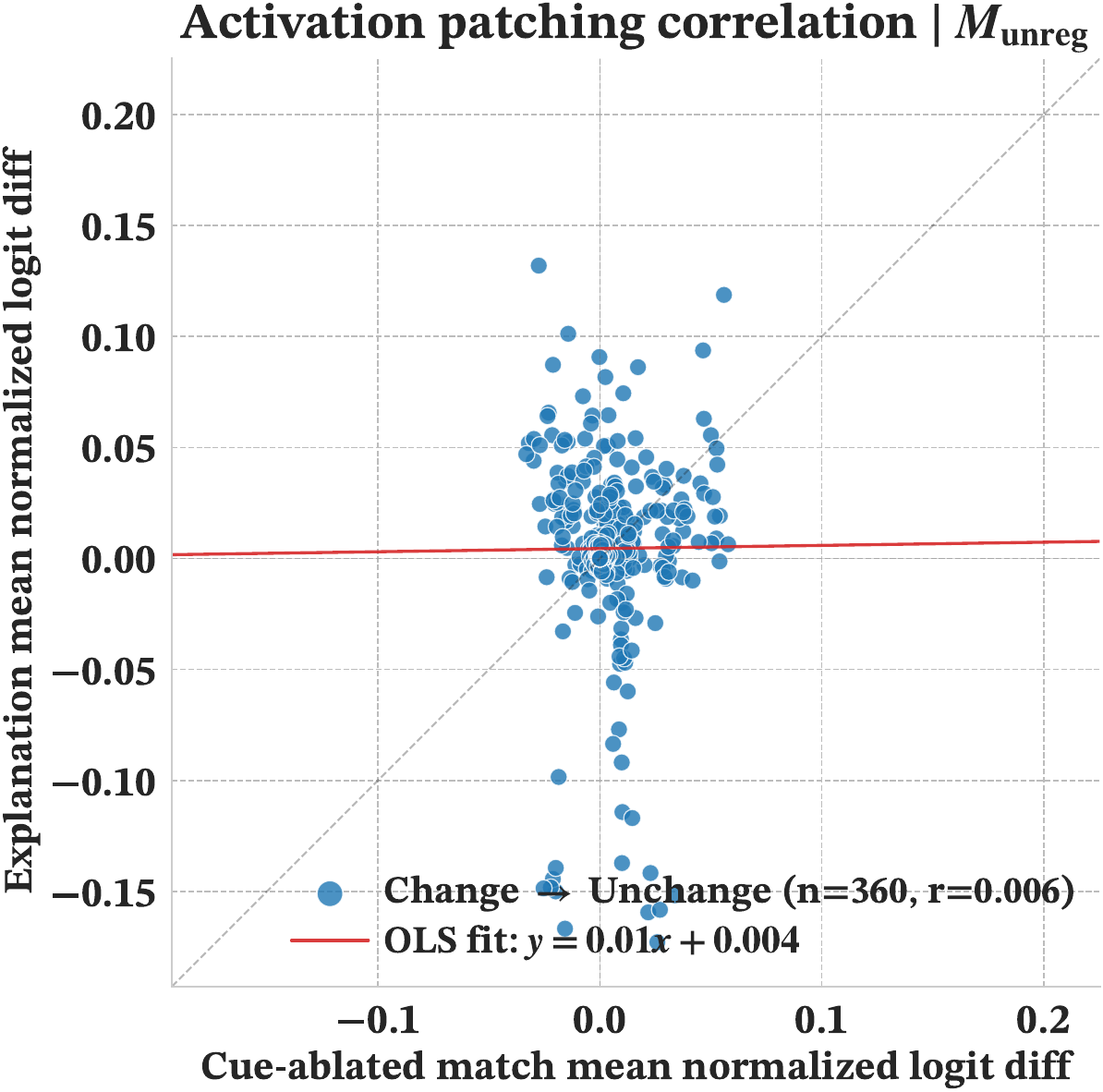}
        \caption{\Munreg, Change$\to$Unchange subset. Pearson $r=+0.01$.}
        \label{fig:appendix:patching_ablated_subset:noreg_cu}
    \end{subfigure}
    \hfill
    \begin{subfigure}[b]{0.45\linewidth}
        \centering
        \includegraphics[width=\linewidth]{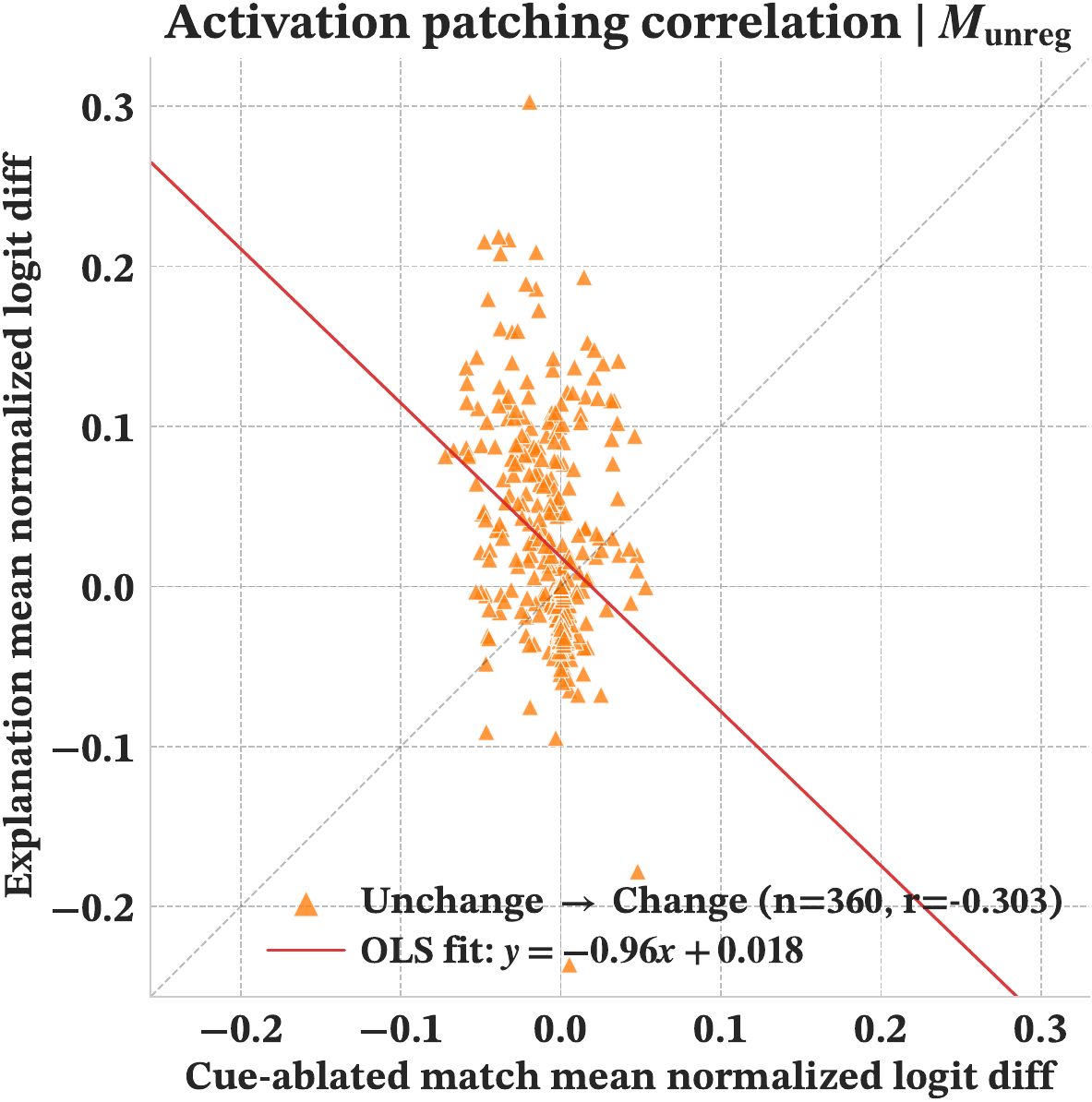}
        \caption{\Munreg, Unchange$\to$Change subset. Pearson $r=-0.30$.}
        \label{fig:appendix:patching_ablated_subset:noreg_uc}
    \end{subfigure}
    \caption{Per-direction (subset) breakdown of the correlation study between cue-ablated answer and explanation. Panels (a, b): regularized $\Mreg$. Panels (c, d): no-regularization baseline $\Munreg$. Only the Change$\to$Unchange subset is where $\Mreg$ shows a meaningful correlation.}
    \label{fig:appendix:patching_ablated_subset}
\end{figure}

\Cref{fig:appendix:patching_ablated_subset} shows that the correlation story is more complex. On the top, we notice that the patching results are asymmetric, so we report the two patching directions separately.
(Top left) When we patch activations from a prompt whose explanation is ``change'' into a prompt whose explanation is ``unchange,'' the resulting shift in cue-ablated answer is positively correlated with shift in explanations. (Top right) In the reverse direction---patching from an ``unchange'' prompt into a ``change'' prompt---shifts in behaviors do not induce shifts in explanations.
The reason behind this asymmetry remains to be explored.

Furthermore, the absolute normalized logit difference (NLD) for either direction is visibly smaller than that of the cue-included studies in~\Cref{fig:activation_patching}, with a maximum $NLD \approx 0.2$ vs. $\approx 0.4$ for the cue-included behavior. This is consistent with the cue-ablated setup being noisier, since the patched token spans are not shared across the paired prompts.

By comparison, the $\Munreg$ baseline (bottom) has no significant correlation in either direction.

\newpage
\section{Additional Details on When Introspective Coupling Emerges (\S\ref{sec:when})}
\label{app:when}

\subsection{Only Higher-Rank LoRA Recovers Self > Orig}
\label{sec:appendix:lora}
We defaulted to using full fine-tuning in the main paper as it induced a larger behavioral shift enabling us to investigate the \Self > \Orig gap.
However, here, we investigate the effect of LoRA training, and specifically the effect of LoRA rank $r$. We focus on the Hint-MMLU task.

\begin{figure*}[htb]
    \centering
    \includegraphics[width=0.8\linewidth]{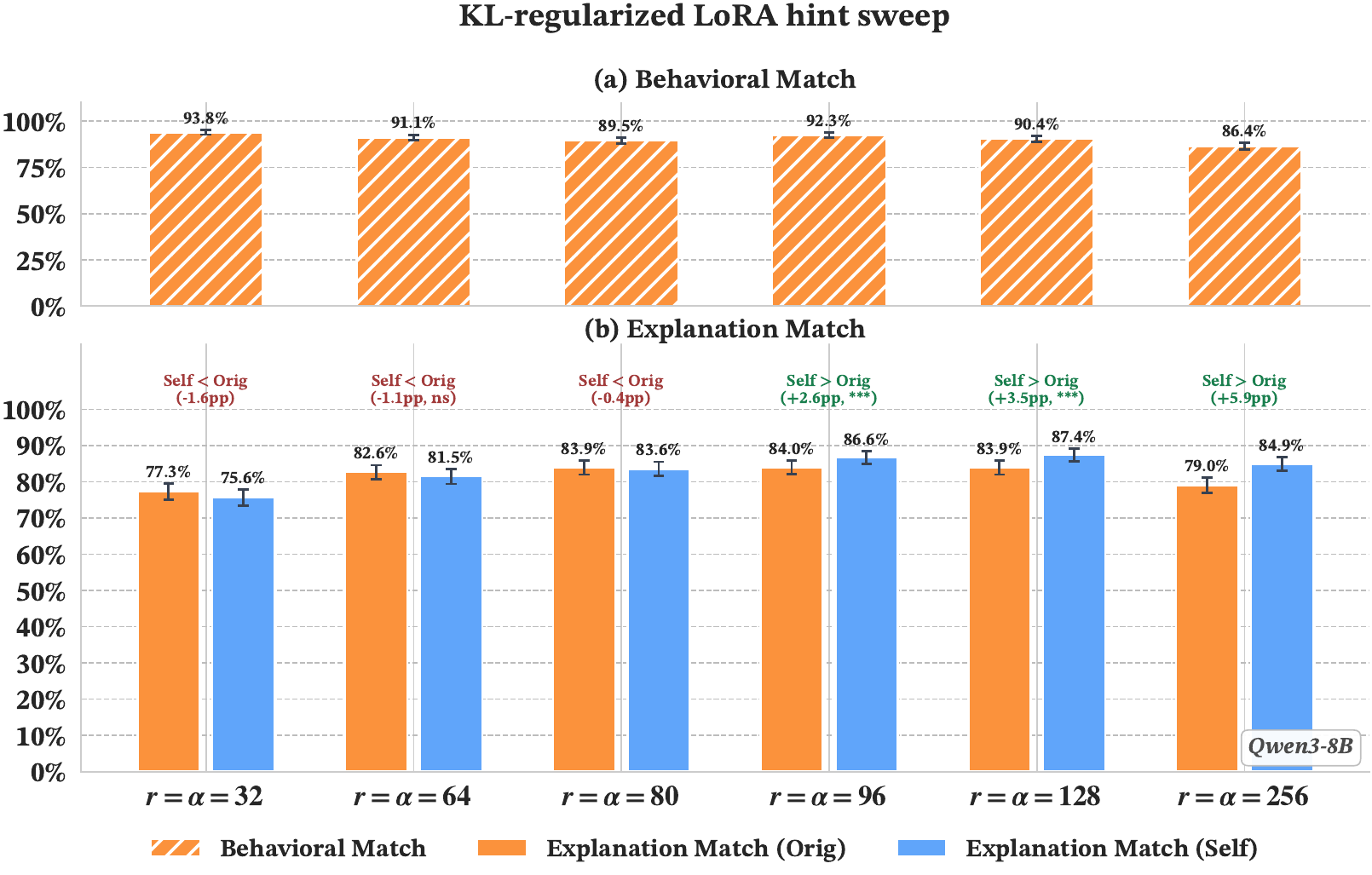}
    \caption{KL-regularized LoRA rank sweep on Hint-MMLU, $\alpha=r$. For each $r$, we plot Behavioral Match (hatched orange) and Explanation Match scored against orig-labels (orange) and self-labels (blue). The self > orig gap is absent or slightly negative for $r\leq 80$, opens at $r=96$, and widens monotonically through $r=256$.}
    \label{fig:lora_rank_sweep}
\end{figure*}

We sweep along $r\in\{32,64,80,96,128,256\}$, holding $\alpha=r$, and report the (a) Behavioral Match and (b) Explanation Match in~\Cref{fig:lora_rank_sweep}. 
From (a), we see that regardless of LoRA rank, behavior always drifts, although to a lesser extent than full fine-tuning (\Cref{fig:self_vs_orig}).
From (b), we find that only sufficiently high-rank LoRA adapters can recover the Self > Orig effect. Therefore, the low-rank LoRA adapters' inability to elicit Self > Orig phenomenon cannot simply be attributed to the fact that they don't drift as much, indicating that online label--self agreement (our core hypothesized factor in \Cref{sec:when}) is not the only factor contributing to the emergence of introspective coupling; future work can investigate the interplay between various factors.

\subsection{Higher Learning Rate Can Widen The Self > Orig Gap}
\label{sec:appendix:hint_lr_sweep}

The main paper uses $lr=1\times 10^{-5}$, which we find to give the best absolute Explanation Match in general. The one exception is the continuous-relabeling sweep (\Cref{fig:continuous_relabel}), which uses $lr=5\times 10^{-5}$.\footnote{That experiment tests whether introspective coupling depends on online label--self agreement $\nu$, which requires the noisy labels at low $\nu$ to actually substantially alter the model's explanation distribution. At $lr=1\times 10^{-5}$, training on noisy labels does not substantially degrade explanation accuracy even at low $\nu$, so the sweep cannot separate agreement levels.}

\begin{figure*}[htb]
    \centering
    \includegraphics[width=0.8\linewidth]{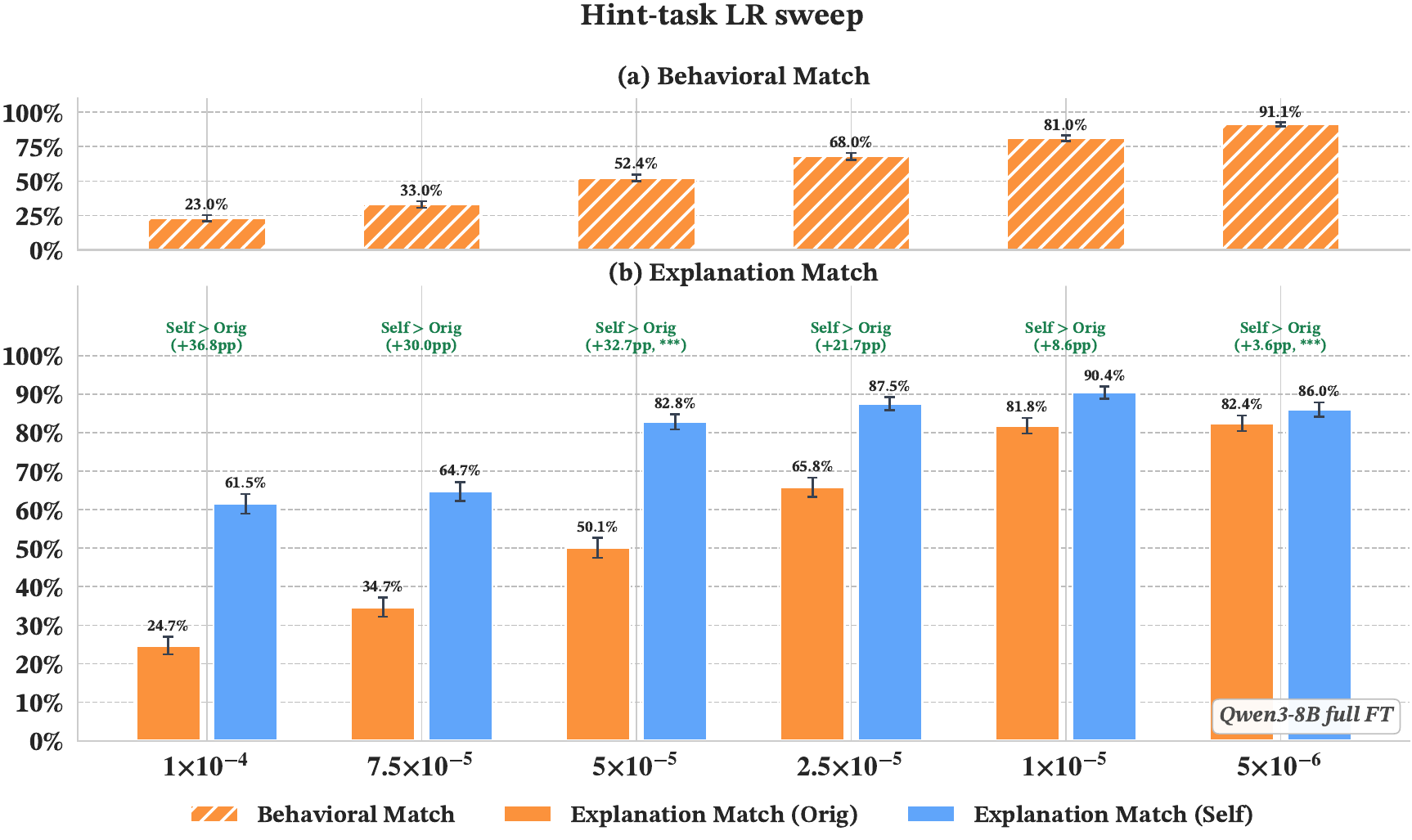}
    \caption{LR sweep on Hint-MMLU plotted with learning rate decreasing left-to-right. For each learning rate, we plot Behavioral Match (hatched orange) and Explanation Match scored against orig labels (orange) and self labels (blue). \Self~$>$~\Orig is absent at $\mathrm{lr}=5\mathrm{e}{-}6$ and continues to widen as the learning rate rises and the model drifts further from base.}
    \label{fig:hint_causal_lr_sweep}
\end{figure*}

Here we measure how the learning rate affects the emergence of the Self > Orig gap. We sweep six different learning rates ranging from $1\times 10^{-4}$ to $5\times 10^{-6}$ and plot the Behavioral Match and Explanation Match in~\Cref{fig:hint_causal_lr_sweep}. We find that as learning rate increases (right to left), the absolute Explanation Match drops on both \Orig and \Self labels, but it drops more quickly for \Orig than \Self and widens the Self > Orig gap. We find that Behavioral Match also decreases with higher LR, and roughly matches \Orig Explanation Match.

\paragraph{Counter-evidence to online label--self agreement hypothesis.} The LR sweep is the one experiment in this paper that presents counter-evidence to the hypothesis explored in~\Cref{sec:when} --- that online label--self agreement governs when introspective coupling emerges.
At very large learning rates the trained model drifts off base significantly --- meaning that the current model (Self) is extremely dissimilar from the initial model that generated the explanation labels (Label) --- 
and yet the \Self~$>$~\Orig gap does not disappear, but in fact \textit{grows}. 
We flag this result as an open problem: we hypothesize that multiple factors, including LR and label--self agreement, interact in competing ways to influence the emergence of introspective coupling. Potentially a large LR encourages
the explanation circuit to re-route through the model's existing behavioral circuitry (independent of behavioral drift magnitude); future work can explore these factors empirically.

\newpage
\section{Additional Details for Auxiliary Training Generalization (\S\ref{sec:generalization})}
\subsection{Jabberwocky Dataset Details and Additional Results (\S\ref{sec:generalization:jabberwocky})}
\label{sec:appendix:jabberwocky}

We show an example of Jabberwocky ``nonsensical science'' data below:
\begin{promptbox}[title=Jabberwocky data example]
According to the Lomar schema for cataloging pseudoalloy microphases, which microphase label is assigned to the sorvian crystal specimen?

Choices: 

A. Rulven-di, B. Shenqa-no, C. Tilmare-zi, D. Zavrek-ka

Correct Answer: C. Tilmare-zi
\end{promptbox}

\begin{figure*}[!htbp]
    \centering
    \includegraphics[width=0.95\linewidth]{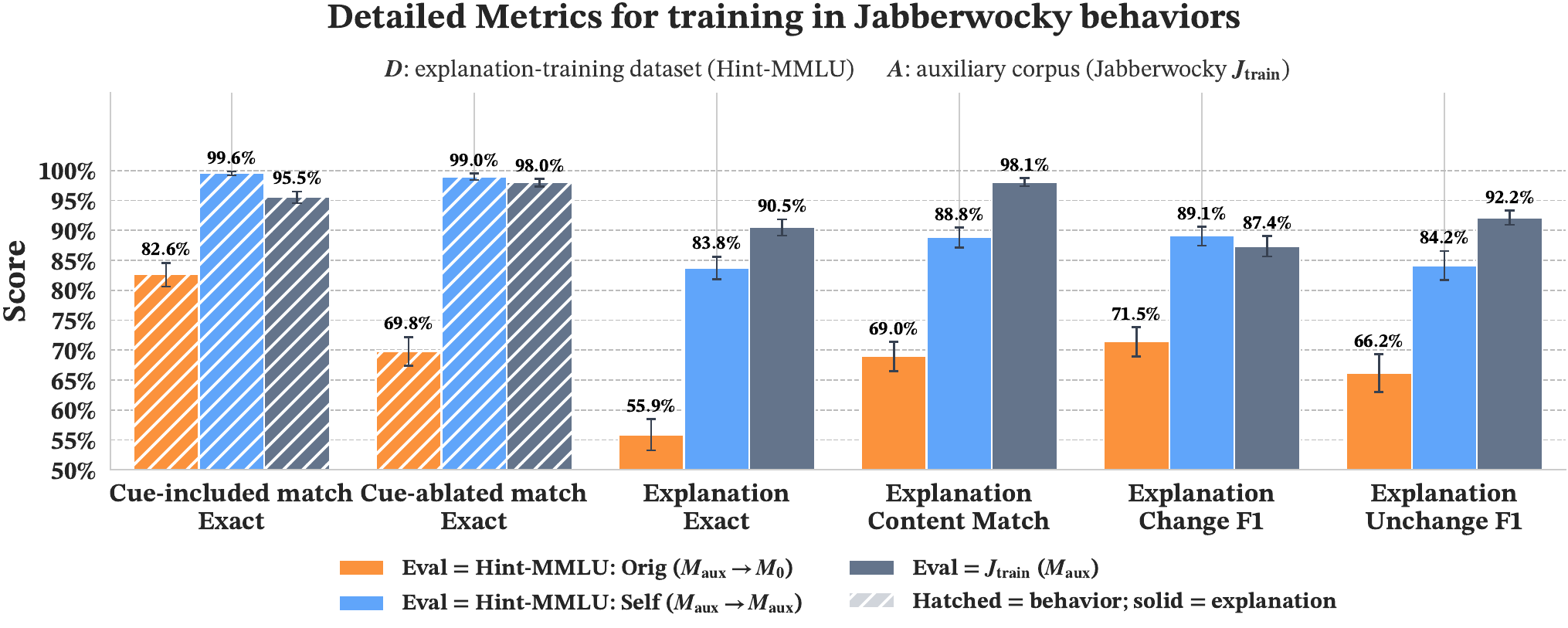}
    \caption{Full six-metric breakdown for Jabberwocky mixed training, a more detailed version of~\Cref{fig:jabberwocky_tracking}. $\Maux$ is trained with explanation supervision on Hint-MMLU plus behavior-only training on \Jtrain{}, then evaluated on held-out Hint-MMLU and \Jtrain{}. Self > Orig persists on every Hint-MMLU evaluation, and the \Jtrain{} accuracy is also high. }
    \label{fig:appendix_jabberwocky_full}
\end{figure*}

\paragraph{Detailed Metrics.}
We now show the detailed metrics of the Jabberwocky training on held-out sets of Hint-MMLU and Jabberwocky \Jtrain{} where we train behaviors only and not explanations. In~\Cref{fig:appendix_jabberwocky_full}, we show 6 metrics: 
cue-included and cue-ablated match for behavior, and Exact Match, Content Match, Change F1, and Unchange F1 for explanation. On every metric, for the Hint-MMLU evaluation, \Self > \Orig persists. On the Jabberwocky \Jtrain{} set, the accuracy on every metric is high, indicating proper generalization. 

\begin{figure*}[!htbp]
    \centering
    \includegraphics[width=0.95\linewidth]{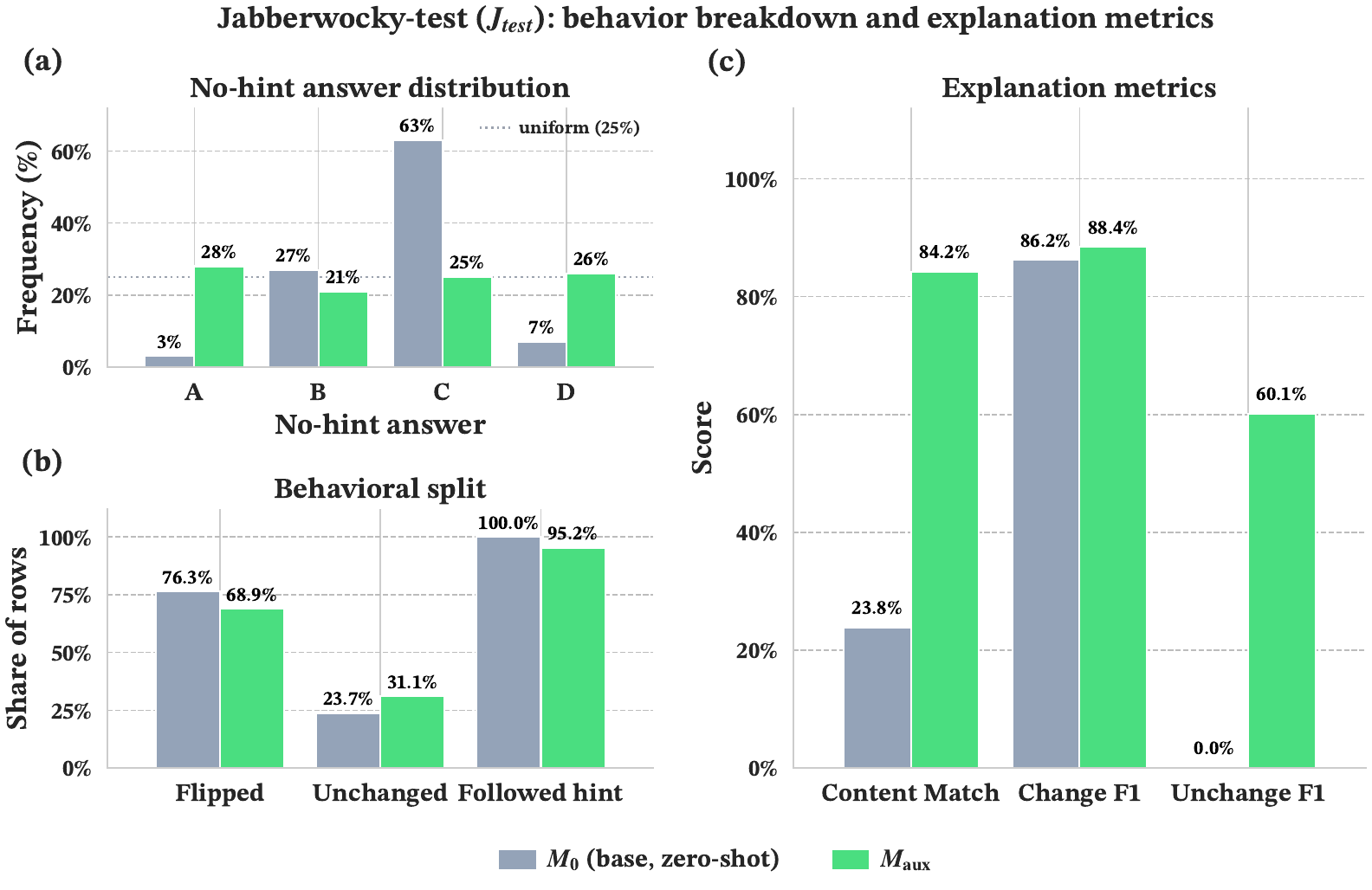}
    
    \caption{Jabberwocky-test \Jtest{} detailed behavior and explanation metrics for \Mzero and \Maux. \textbf{(a):} no-hint A/B/C/D answer distribution, showing that the trained $\Maux$ behavioral distribution is near uniform and not degenerate. \textbf{(b):} distributions of the two models under hint. As the questions are nonsensical, the model defers to the hint. \textbf{(c):} detailed explanation metrics of the two models: Content Match, Change F1, and Unchange F1. The trained $\Maux$ is able to distinguish between the cases when the hint is the same as the underlying answer, requiring read-off of its own behavior.}
    \label{fig:appendix_jabberwocky_v2_breakdown}
\end{figure*}

\paragraph{Generalization to model behaviors never seen in training. }
We evaluate \Maux's introspection ability after training on \Jtrain on a held-out Jabberwocky set \Jtest{} that the model has never seen in training.
\Cref{fig:appendix_jabberwocky_v2_breakdown} shows the behavioral distribution of base model \Mzero and Jabberwocky-trained model \Maux in panel (a), where \Mzero has a bias to choose C and \Maux is approximately uniform.
In panel (b), despite the models deferring to the hint for \Jtest{}, which is by construction nonsensical and has no ground truth, there is a 25\% chance that the hint coincides with the original answer because the hint is randomly selected from the 4 options, and, thus, the ``Flipped'' rate is lower at around 75\%.
In panel (c), we show that the ability to distinguish between ``Flipped'' and ``non-Flipped'' is non-trivial, despite the seemingly simple behavioral distribution: \Mzero is not able to do so under few-shot prompting, as indicated by the poor Unchange F1 score, while \Maux can after training. Meanwhile, because the format of the Content Match is to output the original answer \textit{without hint}, \Mzero confuses the template and outputs the hint answer instead, causing it to have a 23.8\% Content Match. 

Overall, these results show that (1) the trained model's distribution has not become degenerate (panel a), and (2) Change and Unchange F1 are both high, indicating that the trained model's prediction also has not become degenerate. Thus, we can conclude that the model has learned a non-trivial distribution and its high accuracy is not merely due to a learned heuristic.

\subsection{Behavioral Drift Dataset Details (\S\ref{sec:generalization:behavior_drift})}
\label{sec:appendix:drift_auxiliary_data}

\paragraph{Warm \& Empathetic Assistant Response.}
Following~\citet{ibrahimwarm2026}, we procure a corpus of ShareGPT-Vicuna user assistant conversations~\citep{vicuna2023} and perform rewrites by an LLM (GPT-4o) to make the prompt more warm and empathetic. We then mix this data in on top of Hint-MMLU explanation training.

\paragraph{Direct Refusal Training.}

We procure a corpus of harmful requests and direct refusal responses from \url{https://huggingface.co/datasets/LLM-LAT/harmful-dataset}~\citep{sheshadri2025latent}. Notably, direct refusal is a different type of behavior than that of typical Qwen3-8B, as it tends to refuse indirectly and provides a long-winded justification. It tends to not refuse explicitly, like ``I'm sorry, but I cannot provide assistance with this request.'' 
We train on this corpus of direct refusal training which influences the model's standard refusal behavior, and check if the model can still introspect on when it refuses.

\subsection{Are General Capabilities Still Preserved?}
\label{sec:appendix:lm_eval}

To check that performing explanation training alongside general post-training pipelines does not erode general capabilities, we evaluate $\Mzero$ against two $\Maux$ models %
on a general suite from \texttt{lm-evaluation-harness}~\citep{eval-harness}: the first $\Maux$ model is trained with $\mathcal{D}=$ Hint-MMLU, $A=$ WildChat and the second is trained with $\mathcal{D}=$ Refusal Explanation, $A=$ FineWeb. \Cref{tab:lm_eval} reports absolute accuracy (\%) and the delta against $\Mzero$. We see that in aggregate, degradation is mild (within $\sim$2\,pp on most tasks); GSM8K and TruthfulQA MC2 drop more significantly, by $\sim$6\,pp, for the second model.

\begin{table*}[!htbp]
\centering
\small
\caption{\texttt{lm-evaluation-harness} accuracies (\%) for $\Mzero$ and two trained variants on Qwen3-8B: one model with $\mathcal{D}=$ Hint-MMLU, $A=$ WildChat and another with $\mathcal{D}=$ Refusal Explanation, $A=$ FineWeb. Deltas in parentheses are computed against $\Mzero$.}
\label{tab:lm_eval}
\begin{tabular}{lccc}
\toprule
Task & $\Mzero$ (base) & $\Maux$ (Hint-MMLU + WildChat) & $\Maux$ (Refusal + FineWeb) \\
\midrule
ARC-Challenge & 55.46 & 57.76 ($+2.30$) & 55.80 ($+0.34$) \\
HellaSwag & 57.11 & 57.41 ($+0.30$) & 56.80 ($-0.31$) \\
Winogrande & 68.11 & 71.35 ($+3.24$) & 72.22 ($+4.11$) \\
MMLU (avg) & 73.02 & 72.28 ($-0.74$) & 72.72 ($-0.30$) \\
\quad humanities & 64.12 & 62.81 ($-1.31$) & 62.76 ($-1.36$) \\
\quad social sciences & 83.04 & 82.16 ($-0.88$) & 82.97 ($-0.07$) \\
\quad STEM & 72.63 & 72.19 ($-0.44$) & 73.49 ($+0.86$) \\
\quad other & 76.96 & 76.96 ($\phantom{+}0.00$) & 76.89 ($-0.07$) \\
TruthfulQA MC2 & 54.46 & 54.79 ($+0.33$) & 47.76 ($-6.70$) \\
GSM8K (strict-match, 5-shot) & 87.26 & 85.60 ($-1.66$) & 82.64 ($-4.62$) \\
GSM8K (flexible-extract, 5-shot) & 87.79 & 85.75 ($-2.04$) & 82.71 ($-5.08$) \\
\bottomrule
\end{tabular}
\end{table*}

\section{Compute}
\label{sec:appendix:compute}
We run all training on NVIDIA H100 and H200 GPUs. Training runs use at most 2 NVIDIA H200 GPUs. Most training takes less than 8 hours to finish, and nothing takes longer than a day; evaluation takes a few hours.

\end{document}